\documentclass[a4paper,fleqn]{cas-sc}

\usepackage[numbers]{natbib}
\usepackage{adjustbox}
\usepackage{textcomp}
\usepackage{etoolbox,lipsum,cleveref}
\usepackage[font=footnotesize,labelfont=bf]{caption}
\usepackage[font=footnotesize,labelfont=bf]{subcaption}
\usepackage{url}

\usepackage{pifont}% http://ctan.org/pkg/pifont
\usepackage{footnote}
\usepackage[flushleft]{threeparttable}

\usepackage{makecell}

\usepackage{hyperref}

 \usepackage{ragged2e}
\def\tsc#1{\csdef{#1}{\textsc{\lowercase{#1}}\xspace}}
\tsc{WGM}
\tsc{QE}
\tsc{EP}
\tsc{PMS}
\tsc{BEC}
\tsc{DE}
\begin{document}
\let\WriteBookmarks\relax
\def\floatpagepagefraction{1}
\def\textpagefraction{.001}

\shorttitle{OCR for Arabic}

\shortauthors{Mahmoud Kasem}

\title [mode = title]{Advancements and Challenges in Arabic Optical Character Recognition: A Comprehensive Survey}                      


\author[1,3]{Mahmoud SalahEldin Kasem}
%\ead{mahmoud.salah@aun.edu.eg}

\author[1,2]{Mohamed Mahmoud}
\author[1]{Hyun-Soo Kang\corref{cor1}}
\ead{Hyun-Soo Kang, hskang@cbnu.ac.kr}

\address[1]{
Department of Information and Communication Engineering, School of Electrical and Computer Engineering, Chungbuk National University, Cheongju-si 28644, Republic of Korea}
\address[2]{
Information Technology Department, Faculty of Computers and Information, Assiut University, Assiut 71526, Egypt}
\address[3]{
Multimedia Department, Faculty of Computers and Information, Assiut University, Assiut 71526, Egypt}

\cortext[cor1]{Corresponding author}
\cortext[cor2]{Principal corresponding author}

\begin{abstract}
Optical character recognition (OCR) is a vital process that involves the extraction of handwritten or printed text from scanned or printed images, converting it into a format that can be understood and processed by machines. This enables further data processing activities such as searching and editing. The automatic extraction of text through OCR plays a crucial role in digitizing documents, enhancing productivity, improving accessibility, and preserving historical records.
This paper seeks to offer an exhaustive review of contemporary applications, methodologies, and challenges associated with Arabic Optical Character Recognition (OCR). A thorough analysis is conducted on prevailing techniques utilized throughout the OCR process, with a dedicated effort to discern the most efficacious approaches that demonstrate enhanced outcomes. To ensure a thorough evaluation, a meticulous keyword-search methodology is adopted, encompassing a comprehensive analysis of articles relevant to Arabic OCR, including both backward and forward citation reviews.
In addition to presenting cutting-edge techniques and methods, this paper critically identifies research gaps within the realm of Arabic OCR. By highlighting these gaps, we shed light on potential areas for future exploration and development, thereby guiding researchers toward promising avenues in the field of Arabic OCR. The outcomes of this study provide valuable insights for researchers, practitioners, and stakeholders involved in Arabic OCR, ultimately fostering advancements in the field and facilitating the creation of more accurate and efficient OCR systems for the Arabic language.

\end{abstract}

% \includegraphics{figs/grabs.pdf}

% Keywords

\maketitle

\section{Introduction}

Deep learning, a notable subfield of machine learning, has experienced extensive utilization across diverse industries. In the realm of computer vision, deep learning algorithms have been successfully employed for critical tasks such as object detection, image classification, and video analysis. Natural Language Processing (NLP) has also witnessed the application of deep learning models in various domains, including text classification, Question Answering, sentiment analysis, sentence similarity \cite{mamdouh2020sentence,abdallah2023exploring,abdallah2023generator,abdallah2023amurd}, machine translation, speech recognition \cite{chorowski2015attention}, and table detection and recognition \cite{prasad2020cascadetabnet,abdallah2022tncr,kasem2022deep}. The healthcare industry has leveraged deep learning for multiple applications, including diagnosis, treatment planning, drug discovery, and medical imaging analysis \cite{nie2015disease,abdallah2020automated,mahmoud2023ganmasker,abdimanap2022enhancing}. Robotics extensively relies on deep learning techniques for autonomous navigation, object recognition, and robotic control. Additionally, deep learning has demonstrated its efficacy in handwritten recognition for numerous languages \cite{mahmoud2014khatt,nurseitov2021handwritten,toiganbayeva2022kohtd,nurseitov2021classification,abdallah2020attention}. In the Internet of Things (IoT) landscape, deep learning finds its application in intrusion detection \cite{mahmoud2022ae,xu2021improving,akkad2023information}. The finance industry has also embraced deep learning in areas such as fraud detection, algorithmic trading, and risk management. Furthermore, deep learning is increasingly finding use cases in the gaming industry, contributing to advancements in game playing, decision-making, customer segmentation, personalized recommendations, and sentiment analysis. The transportation industry has witnessed a profound impact through deep learning, with applications spanning autonomous vehicles, traffic prediction, and route optimization. Finally, the energy industry envisions potential applications of deep learning in predictive maintenance, energy consumption prediction \cite{waschneck2018optimization,hamada2021neural}, and equipment malfunction detection. The aforementioned examples only scratch the surface of the diverse applications of deep learning across various fields, underscoring its immense potential for further growth and development.

Optical Character Recognition (OCR) represents a transformative technology pivotal in the extraction of both handwritten and printed text from scanned or printed images. Its fundamental role lies in the conversion of such text into a machine-readable format, facilitating subsequent data processing activities. OCR has revolutionized the way documents are digitized, improving productivity, accessibility, and preservation of historical records. By automating the extraction of text from physical documents, OCR enables efficient searching, editing, and analysis of textual information\cite{alhomed2018survey}.

\subsection{Challenges in Arabic OCR}

Arabic OCR presents several unique challenges that distinguish it from OCR for other languages. This section explores these challenges in detail, highlighting the complexities associated with Arabic text and the implications for OCR accuracy and performance. Understanding these challenges is crucial for developing effective OCR solutions tailored to the Arabic language.
\begin{itemize}

    \item Complex Morphology: Arabic features a rich morphological structure, characterized by root-based word formation, extensive use of diacritical marks (i.e., vowel markers and other orthographic symbols), and various forms of ligatures and connecting letters. These complexities pose challenges for segmentation, character recognition, and word-level analysis during the OCR process.
    
    \item Contextual Variations: Arabic text exhibits contextual variations that impact character shapes, particularly due to the presence of initial, medial, final, and isolated forms of letters. Accurately recognizing and disambiguating these contextual variants is essential for maintaining OCR accuracy.

    \item Cursive Writing Style: Arabic is commonly written in a cursive style, where letters are connected, resulting in overlapping strokes. This cursive nature makes it challenging to separate individual characters accurately, affecting character segmentation and recognition algorithms.

    \item Diacritic Marks: Arabic utilizes diacritical marks to indicate vowels and other phonetic information. However, diacritical marks are often omitted in everyday writing or handwriting, making it difficult to accurately reconstruct the original text during OCR.

    \item Ligatures and Shaping: Arabic includes ligatures and contextual shaping, where letters change their shapes based on their position within a word. Proper recognition and interpretation of ligatures and shaping are critical for accurate OCR output.

    \item Limited Availability of Labeled Datasets: Building robust Arabic OCR systems requires large, high-quality, and diverse labeled datasets. However, the availability of such datasets for training and evaluation purposes is limited compared to other languages, which poses challenges for developing and benchmarking Arabic OCR algorithms.

    \item Arabic Dialects and Variations: Arabic is spoken across different regions, leading to dialectal variations in written form. OCR systems must account for these variations to ensure accurate recognition and understanding of Arabic text from different dialectical sources.

\end{itemize}

In the subsequent sections of this paper, we will explore the state-of-the-art techniques and methodologies employed to tackle these challenges and improve the performance of Arabic OCR systems. By addressing these challenges head-on, researchers aim to develop more robust, accurate, and efficient OCR solutions tailored specifically to the complexities of Arabic language processing.

\subsection{Classification of OCR Systems}

Various classifications of OCR systems are contingent upon the language and writing style found in the images. Documents may incorporate handwritten, printed, or scanned content and may involve one or multiple languages. Consequently, OCR systems can be categorized as unilingual or multilingual, predominantly based on language support. A unilingual OCR system is specifically designed to identify a single language, with the Arabic OCR model serving as an illustration of such a system. Conversely, certain OCR systems possess the capability to perform recognition and extraction tasks across multiple languages, earning them the designation of multilingual OCR systems.

Optical Character Recognition (OCR) systems can be broadly classified into two categories: offline OCR systems and online OCR systems. Offline OCR systems are designed to process scanned, printed, and handwritten documents \cite{djaghbellou2021survey}. These systems provide a spectrum of online services tailored to diverse applications, encompassing functionalities such as sorting mails, reading the bank cheques, verification of signatures, processing of utility bills, and applications within the insurance sector. Moreover, offline OCR systems play a crucial role in enhancing accessibility for blind or illiterate individuals by utilizing digital pens to convert text into audio format. In addition, online recognition systems find extensive implementation in diverse domains such as number-plate recognition\cite{islam2017survey}.

On the other hand, online OCR systems are specifically designed to receive and process real-time input images. These systems are capable of handling dynamic, on-the-fly recognition tasks. In contrast, offline recognition in OCR often involves the use of multiple models with different datasets and algorithms to achieve higher levels of recognition accuracy. By employing a variety of models and datasets, offline OCR systems aim to optimize performance and enhance the overall accuracy of text recognition\cite{rashid2022scrutinization}.

\subsection{Applications}
OCR systems have become increasingly prevalent across various industries, offering expedited and precise workflows. The utilization of OCR in digitization processes has proven instrumental. A Singh\cite{singh2012survey} presents an inclusive survey encompassing the application of OCR and conduct experiments pertaining to select use cases. The discussed OCR applications include:
\begin{itemize}
    \item Invoice Imaging: Widely employed in numerous businesses to facilitate the efficient tracking of crucial business records.
    \item Banking: OCR finds extensive use in banking services. For instance, the scanning and rapid transfer of check payments significantly expedite the processing time.
    \item Captcha: Captcha is employed as a security measure in systems. It involves distorting an image containing a combination of letters, numbers, or both. Humans can easily decipher such captchas, while it poses a challenge for average computer programs.
    \item Legal Industry: OCR is employed to documents digitization, streamlining operations within the legal sector.
    \item Automatic Number Recognition: Surveillance systems utilize OCR for tracking vehicle records by capturing and recognizing number plates. OCR aids in the accurate extraction of numbers and characters from the plates.
    \item Handwriting Recognition: The process entails extracting text from handwritten photographs and documents. The OCR model undergoes learning to proficiently identify diverse fonts also different languages, contributing to an elevated level of recognition accuracy.
    \item Scanned Receipts: Challenges arise when extracting information from scanned receipts due to layout variations, noise, and distortion \cite{antoniosurvey}. Overcoming these obstacles using OCR is crucial for efficient data extraction.
    \item Healthcare: OCR plays a pivotal role in processing a wide range of patient data, including forms, reports.
\end{itemize}

\subsection{Procedural Overview of Arabic OCR Systems}
\begin{figure}[h!]
    
    \includegraphics[width=\textwidth,height=0.08\textwidth]{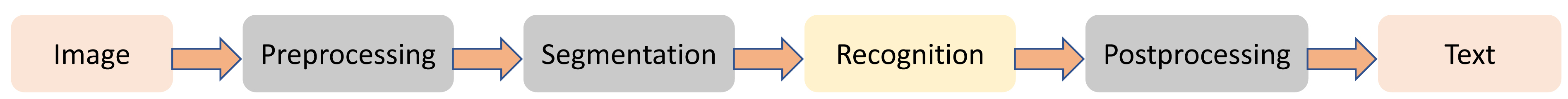}
    
\caption{Brief overview of OCR process} 
\label{fig: Brief overview of OCR process}
\end{figure}

Arabic OCR systems encompass a series of intricate and well-defined procedures, as depicted in Figure \ref{fig: Brief overview of OCR process}, which provides a succinct overview of the specific steps that must be followed. The initial phase involves the preprocessing of the image to enhance its quality and optimize the recognition process. The preprocessing phase involves a series of operations, including skewing correction, reduction of noise, and enhancement of the contrast. Subsequently, the area contains text in the image undergoes meticulous segmentation into individual units, comprising characters or Texts. These characters that are segmented are subjected to recognition, with the most suitable counterpart determined by utilizing a comprehensive dataset of characters known. The text that is recognized subsequently progresses through additional processing stages, focusing on error correction and overall accuracy improvement. Ultimately, the output materializes as a machine-readable text document, granting software applications the capability to seamlessly edit, search, and analyze the content.

\subsection{Goals and Outlines}

This paper provides a detailed exploration of OCR with a specific focus on Arabic language processing. Arabic, as a complex and highly contextual language, presents unique challenges that necessitate specialized OCR techniques. Understanding and accurately processing Arabic text is vital for a wide range of applications, including document digitization, text analysis, information retrieval, and language preservation.

The objective of this paper is to present an in-depth examination of contemporary applications, methodologies, and challenges in Arabic OCR, encompassing cutting-edge advancements in the field. By systematically reviewing current methods employed at every step of the process of the OCR, we aim to differentiate the most effective approaches that yield improved results. This survey follows a rigorous keyword-search methodology, Integrating a thorough analysis of articles pertaining to Arabic OCR, encompassing reviews of both forward and backward citations., to ensure a thorough evaluation of the literature.

Furthermore, this paper goes beyond a mere review of existing techniques and methods. It identifies research gaps and highlights areas for future exploration and development in Arabic OCR. By pinpointing these gaps, we aim to inspire researchers to address the remaining challenges and pave the way for advancements in the field. The outcomes of this study serve as a valuable resource for researchers, practitioners, and stakeholders involved in Arabic OCR, directing them towards promising paths and cultivating the development of OCR systems for the Arabic language that are both more accurate and efficient.

The subsequent sections of this paper are structured as follows: Section 2 provides an in-depth examination of the datasets accessible for evaluating Arabic OCR systems. In Section 3, We offer a thorough examination of the current literature on every stage of Arabic OCR, highlighting significant research trends and advancements. In conclusion, we summarize the key findings from this survey, emphasizing research gaps.

\section{Databases}

\begin{figure}[h!]
    \centering
    \begin{subfigure}{0.50\columnwidth}
        \includegraphics[width=\textwidth,height=0.4\textheight,keepaspectratio]{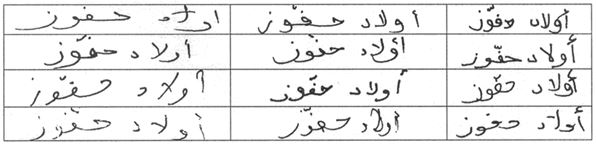}

        \caption{IFN/ENIT}
        %\label{fig:example4}
    \end{subfigure}
    \begin{subfigure}{0.49\columnwidth}
        \includegraphics[width=\textwidth,height=0.4\textheight,keepaspectratio]{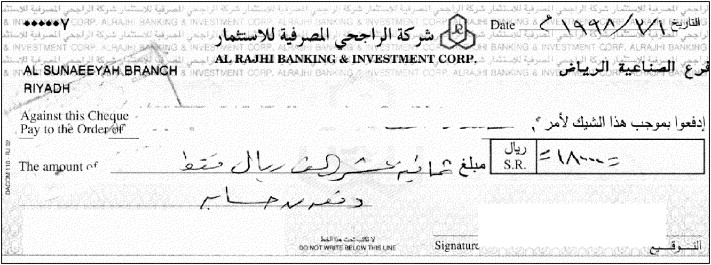}

        \caption{cheque}
        %\label{fig:example4}
    \end{subfigure}
    \begin{subfigure}{0.50\columnwidth}
        \includegraphics[width=\textwidth,height=0.3\textheight,keepaspectratio]{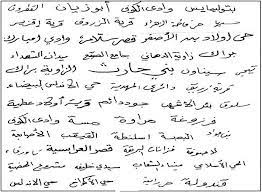}

        \caption{AHDB}
        %\label{fig:example4}
    \end{subfigure}
    \begin{subfigure}{0.49\columnwidth}
    \includegraphics[width=\textwidth,height=0.3\textheight,keepaspectratio]{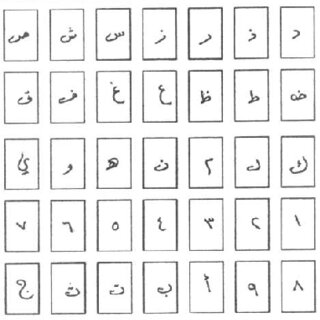}
        \caption{ Forms}
        %\label{fig:example4}
    \end{subfigure}

\caption{Examples of images in IFN/ENIT, Cheque, AHDB, and Forms. } 
\label{fig:DatasetEx1} 
\end{figure}

\begin{figure}[h!]
    \centering
    \begin{subfigure}{0.50\columnwidth}
        \includegraphics[width=\textwidth,height=0.4\textheight,keepaspectratio]{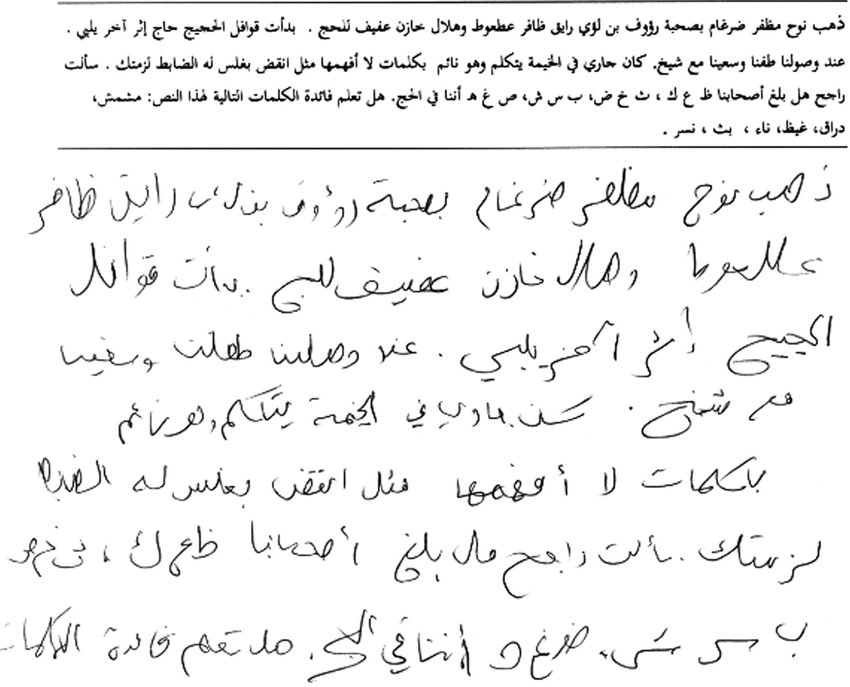}

        \caption{KHATT}
        %\label{fig:example4}
    \end{subfigure}
    \begin{subfigure}{0.49\columnwidth}
        \includegraphics[width=\textwidth,height=0.4\textheight,keepaspectratio]{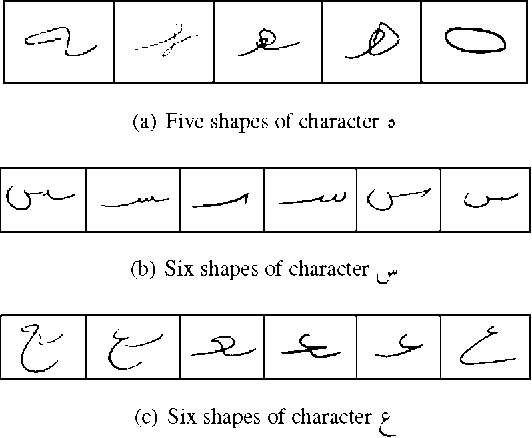}

        \caption{HACDB}
        %\label{fig:example4}
    \end{subfigure}
    \begin{subfigure}{0.50\columnwidth}
        \includegraphics[width=\textwidth,height=0.3\textheight,keepaspectratio]{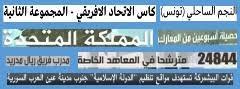}

        \caption{ACTIV}
        %\label{fig:example4}
    \end{subfigure}
    \begin{subfigure}{0.49\columnwidth}
    \includegraphics[width=\textwidth,height=0.3\textheight,keepaspectratio]{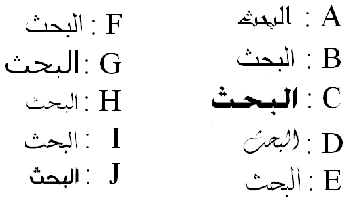}
        \caption{ APTI}
        %\label{fig:example4}
    \end{subfigure}

\caption{Examples of images in KHATT, HACDB, ACTIV, and APTI. } 
\label{fig:DatasetEx2} 
\end{figure}

Dataset plays a pivotal role in the validation of OCR systems, serving as a critical component to assess the accuracy of OCR results. Particularly in the case of the Arabic language, several challenges arise due to its cursive nature, the presence of diacritics, varying writing styles that alter the overall shape of each word, fluctuations in the size of the text, and related factors. Additionally, a compilation of an Arabic database is hindered by the tied availability of resources associated with the Arabic language. In prior works \cite{al2020review,ahmed2019handwritten}, researchers have shared commonly utilized datasets, as showcased in Table \ref{tab:arabicdatasets}, encompassing Arabic, Urdu, and Persian languages, including both publicly accessible and restricted datasets. Also, Figure \ref{fig:DatasetEx1},\ref{fig:DatasetEx2},\ref{fig:DatasetEx3} illustrates examples of each dataset.
\begin{table}[h!]
\caption{comprehensive list of available datasets, accompanied by their corresponding statistics, dataset types, and modes of accessibility}
\begin{adjustbox}{width=1\textwidth}
\begin{tabular}{|c|c|c|c|} 
    \hline
      Dataset & Size & Content & accessibility  \\ \hline
      IFN/ENIT \cite{pechwitz2002ifn}         & 115,000 words \& 212,000 characters  &     Handwritten words     & Public  \\ \hline
      AHDB\cite{al2002data}     & 30,000 words  &     Handwritten words \& digits     & Private \\ \hline
      Cheque\cite{al2003databases}       & 29,498 subwords \& 15,148 digits  &    Handwritten subwords \& digits      & Private \\ \hline
      Forms\cite{asiri2005automatic}  &  15,800 characters \& 500 writers  &  Handwritten characters     & Private \\ \hline
      UPTI\cite{sabbour2013segmentation} & 10,000 lines  &     Printed text lines      & Public  \\ \hline
      Numeral\cite{awaidah2009multiple}       & 21,120 digits \& 44 writers  &     Handwritten digits       & Public \\ \hline
      HACDB\cite{lawgali2013hacdb}            & 6600 characters \& 50 writers  &     Handwritten characters        & Public \\ \hline

      AHDBase\cite{el2007two} & 70,000 digits \& 700 writers  &     Handwritten digits        & Public \\ \hline
      HODA\cite{khosravi2007introducing} & 102,352 digits   &     Handwritten digits        & Public \\ \hline
      APTI\cite{slimane2009database}              & 113,284 words \& 648,280 characters &      Printed words      & Public \\ \hline
      KHATT\cite{mahmoud2014khatt}      & 9327 lines, 165,890 words \& 589,924 characters &      Handwritten text lines      & Public \\ \hline
      ACTIV\cite{zayene2015dataset}            & 4824 lines \& 21,520 words &     Embedded text lines       & Public \\ \hline
      ALIF\cite{yousfi2015alif}         &   1804 words \& 89,819 characters  &  Embedded text lines    & Needs request \\ \hline
      Digital Jawi\cite{saddami2015database}         &  168 words \& 1524 characters   &    Jawi paleography images     & Public  \\ \hline
      SmartATID\cite{chabchoub2016smartatid}       &  9088 pages   & Printed \& handwritten pages & Public \\ \hline
      Printed PAW\cite{bataineh2017printed}      & 415,280 unique words \& 550,000 sub words & Printed subwords  & Needs request  \\ \hline
      Degraded historical\cite{sulaiman2017database}            &  10 handwritten images \& 10 printed images    &  Handwritten documents    & Public  \\ \hline 
      ACTIV2\cite{zayene2018open}         & 10,415 text images &   Embedded words  & Public  \\ \hline 
      QTID\cite{badry2018qtid}         &  309,720 words \& 249,428 characters    &    Synthetic words    & Private  \\ \hline 
      KAFD\cite{luqman2014kafd}              & 28,767 pages \& 644,006 lines &    Printed pages \& lines     & Public  \\ \hline 
      AHDB/FTR\cite{ramdan2013arabic} &   497 images      & Handwritten Text Images &  Public\\ \hline

      ADBase \& MADBase \footnotemark[1]  &   70,000 digits \&  700 writers      & Handwritten Digits &  Public\\ \hline

      AHT2D \cite{ouali2023augmented}\footnotemark[2]   &  -   & Handwritten Text &  paywall \\ \hline

      AHWD\cite{alzrrog2022deep}  & 21,357 words  &     Handwritten words     &  Private \\ \hline

     %\footnote{\url{https://datacenter.aucegypt.edu/shazeem}}
     %\footnote{\url{https://ieee-dataport.org/documents/aht2d-dataset\#files/}}
     %\multicolumn{4}{|c|}{ $^1$\url{https://datacenter.aucegypt.edu/shazeem/} }\\ 
     %\multicolumn{4}{|c|}{$^2$\url{https://ieee-dataport.org/documents/aht2d-dataset\#files/} }\\  \hline

\end{tabular}
\end{adjustbox}
\label{tab:arabicdatasets}
\end{table}

%%%%%%%%%%%%%%%%%%%%%%%%%%%%%%%%%%%%%%%%%%%%%%%%%%%%%%%%%%%%%%%%%%%%%%%%%%%%%%%%%%%%%%%

%%%%%%%%%%%%%%%%%%%%%%%%%%%%%%%%%%%%%%%%%%%%%%%%%%%%%%%%%%%%%%%%%%%%%%%%%%%%%%%%%%%%%%%
\begin{figure}[h!]
    \centering
    \begin{subfigure}{0.50\columnwidth}
        \includegraphics[width=\textwidth,height=0.4\textheight,keepaspectratio]{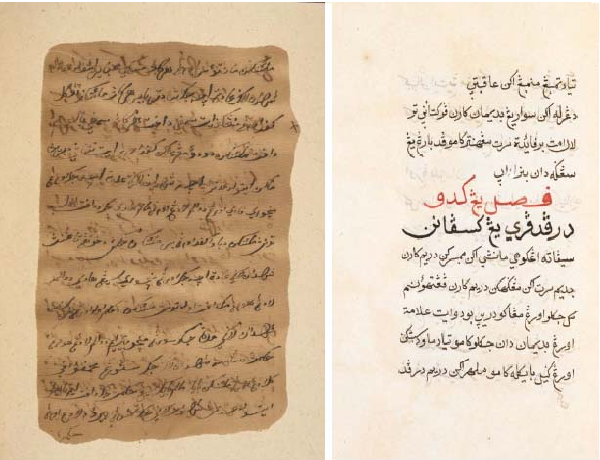}

        \caption{Digital Jawi}
        %\label{fig:example4}
    \end{subfigure}
    \begin{subfigure}{0.49\columnwidth}
        \includegraphics[width=\textwidth,height=0.4\textheight,keepaspectratio]{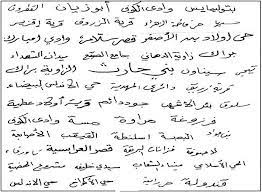}

        \caption{AHDB/FTR}
        %\label{fig:example4}
    \end{subfigure}
    \begin{subfigure}{0.50\columnwidth}
        \includegraphics[width=\textwidth,height=0.3\textheight,keepaspectratio]{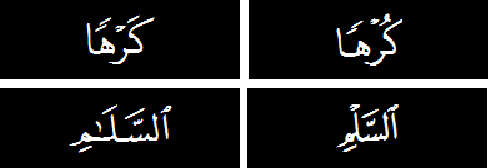}

        \caption{QTID}
        %\label{fig:example4}
    \end{subfigure}
    \begin{subfigure}{0.49\columnwidth}
    \includegraphics[width=\textwidth,height=0.3\textheight,keepaspectratio]{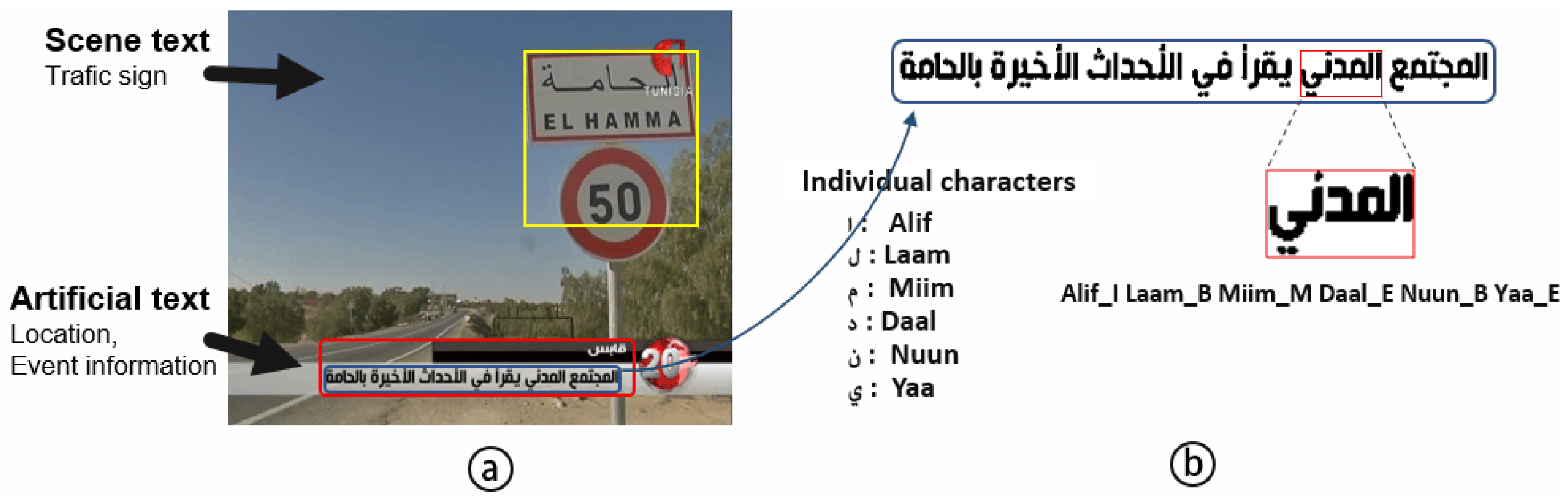}
        \caption{ ACTIV2}
        %\label{fig:example4}
    \end{subfigure}

    \begin{subfigure}{0.50\columnwidth}
        \includegraphics[width=\textwidth,height=0.3\textheight,keepaspectratio]{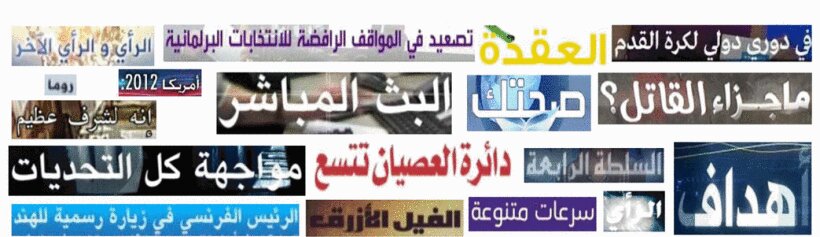}

        \caption{ALIF}
        %\label{fig:example4}
    \end{subfigure}
    \begin{subfigure}{0.49\columnwidth}
    \includegraphics[width=\textwidth,height=0.3\textheight,keepaspectratio]{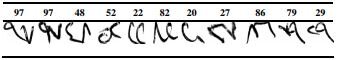}
        \caption{Numeral}
        %\label{fig:example4}
    \end{subfigure}

\caption{Examples of images in Digital Jawi, AHDB/FTR, QTID, ACTIV2, ALIF, and Numeral. } 
\label{fig:DatasetEx3} 
\end{figure} 

%%%%%%%%%%%%%%%%%%%%%%%%%%%%%%%%%%%%%%%%%%%%%%%%%%%%%%%%%%%%%%%%%%%%%%%%%%%%%%%%%%%%%%%

\begin{figure}[h!]
    \centering
    \begin{subfigure}{0.7\columnwidth}
        \includegraphics[width=\textwidth,height=\textheight,keepaspectratio]{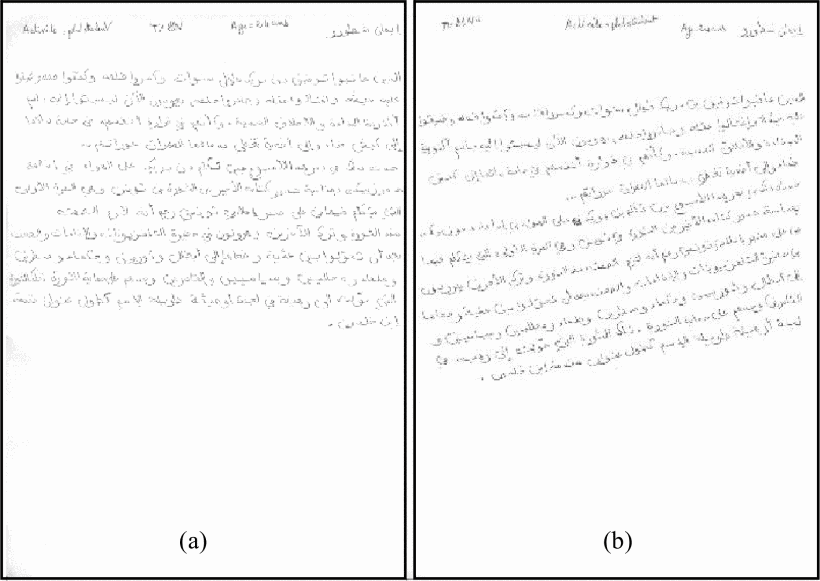}

        \caption{SmartATID}
        %\label{fig:example4}
    \end{subfigure}
    \begin{subfigure}{0.25\columnwidth}
        \centering
        \includegraphics[width=\textwidth,height=0.4\textheight,keepaspectratio]{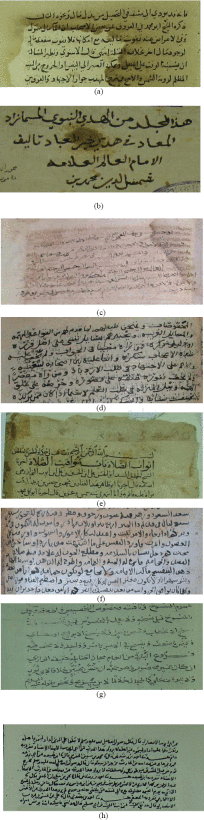}

        \caption{Degraded historical}
        %\label{fig:example4}
    \end{subfigure}

     \begin{subfigure}{0.40\columnwidth}
        \centering
        \includegraphics[width=\textwidth,height=0.3\textheight,keepaspectratio]{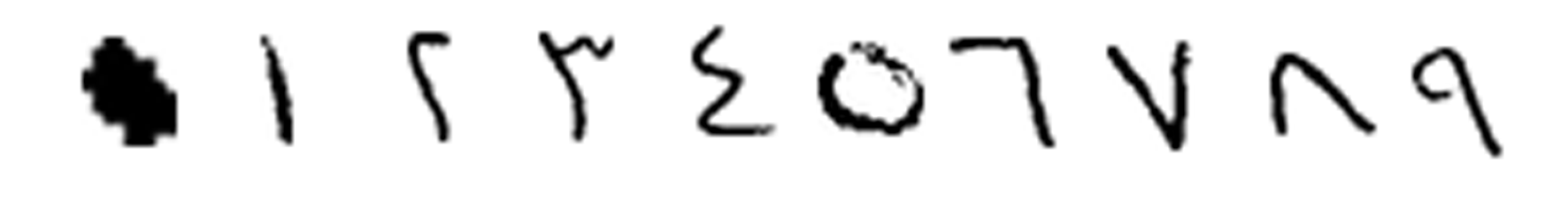}

        \caption{MADBase}
        %\label{fig:example4}
    \end{subfigure}
     \begin{subfigure}{0.41\columnwidth}
        \includegraphics[width=\textwidth,height=0.2\textheight,keepaspectratio]{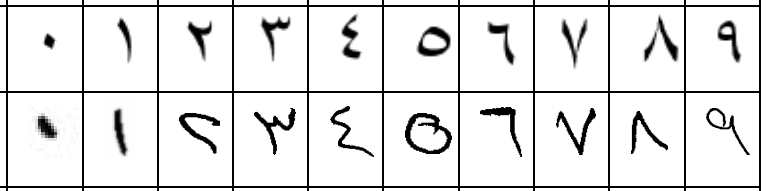}

        \caption{AHDBase}
        %\label{fig:example4}
    \end{subfigure}

    \begin{subfigure}{0.41\columnwidth}
        \includegraphics[width=\textwidth,height=0.2\textheight,keepaspectratio]{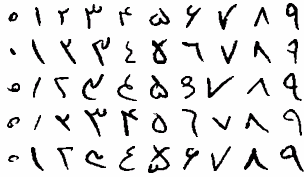}

        \caption{HODA Database}
        %\label{fig:example4}
    \end{subfigure}

\caption{Examples of images in SmartATID, Degraded historical, MADBase, AHDBase, and HODA Database. } 
\label{fig:DatasetEx3} 
\end{figure} 

\subsection{Handwritten Text}
Arabic and Urdu exhibit numerous symmetrics in terms of their writing styles and cursive nature. Both languages adhere to the right-to-left writing direction, while Urdu comprises approximately 39 to 40 letters, with Arabic possessing a slightly smaller character set. Moreover, Urdu extensively incorporates vocabulary borrowed from Arabic, accounting for nearly 30\% of its lexicon. Given these commonalities, datasets and trained models developed for either language are often utilized interchangeably due to their shared characteristics. This cross-usage of resources is particularly beneficial, as it enables leveraging existing data and models, leading to synergistic advancements in both Urdu and Arabic OCR domains.

Datasets for Urdu and Arabic languages are available, showcasing the efforts of researchers in compiling valuable resources. Notably, in \cite{bhatti2023recognition}, a dataset featuring handwritten Urdu numerals was presented. Additionally, \cite{ahmed2019handwritten} introduced the Urdu Nastaliq Handwritten Dataset (UNHD), consisting of handwritten samples contributed by 500 writers on A4-size paper. Furthermore, Khosrobeigi et al. \cite{khosrobeigi2022persian} presented a dataset for the Persian language, which was compiled from various Persian-language news websites, offering a valuable resource for research and development in the field.

Alghamdi and Teahan \cite{alghamdi2018printed} extensively discuss the prevalent datasets commonly employed for the training and assessment of OCR systems focusing on printed form of the Arabic script. They highlight notable datasets, which incorporates datasets such as the Arabic handwritten dataset IFN/ENIT and the "Handwriting Arabic Corpus", and RIMES dataset, encompassing an extensive array of both printed and handwritten documents, this compilation represents a significant resource. The authors deliver a comprehensive overview of these accessible datasets, underscoring the paramount importance of employing high-quality datasets to augment the precision of OCR systems.

\footnotetext[1]{\url{https://datacenter.aucegypt.edu/shazeem}}
\footnotetext[2]{\url{https://ieee-dataport.org/documents/aht2d-dataset\#files/}}

\subsection{Printed Arabic}
Arabic OCR involves the utilization of printed, handwritten, and historical documents. In the context of processing printed Arabic documents, H Bouressace\cite{bouressace2022review} introduces a range of methodologies like bottom-up, top-down, and crossbred approaches. This study extensively examines the various phases involved in the OCR pipeline, including preprocessing, segmentation, feature extraction, and classification. Additionally, A Qaroush\cite{qaroush2022efficient} proposes an effective algorithm for word and character segmentation in printed Arabic documents that is independent of font variations. The algorithm incorporates profile projection for font-independent techniques, while the Inter-quartile Range is leveraged for the segmentation of words. Furthermore, two distinct methods are employed for character segmentation: the holistic approach, which involves a no-segmentation methodology, and the analytical method, characterized by an approach that is based on segmentation.

MA Al Ghamdi\cite{al2022novel} addresses the challenges in Optical Character Recognition (OCR) specific to the Arabic language, driven by the increasing need to digitize Arabic content on the internet. Their focus lies in developing an effective printed Arabic OCR system, divided into four key stages: pre-processing, feature extraction, character segmentation, and classification. Unlike conventional Arabic OCR systems, the unique approach here involves conducting feature extraction before character segmentation. The pre-processing stage incorporates a novel thinning algorithm to generate skeletons for Arabic text images. The feature extraction stage introduces a novel chain code representation technique utilizing an agent-based model for non-dotted Arabic text images. This, in turn, informs a character segmentation technique for breaking down connected Arabic words into individual characters. For classification, the prediction by partial matching (PPM) compression-based method is employed. Experimental evaluation on a public dataset demonstrates the system's accuracy, achieving a notable 77.3\% accuracy for paragraph-based text images.

S Majumdar\cite{majumdar2022recognizing} tackled the intricate challenge of attributing digitized handwriting in a document to multiple scribes, a task characterized by high dimensionality. The dissimilarity in unique handwriting styles arises from a combination of factors, including character size, stroke width, loops, ductus, slant angles, and cursive ligatures. Prior research utilizing labeled data and techniques such as Hidden Markov models, support vector machines, and semi-supervised recurrent neural networks has shown varying degrees of success. In this investigation, the authors achieved successful hand shift detection in a historical manuscript by employing fuzzy soft clustering in conjunction with linear principal component analysis. This notable advancement highlights the effective application of unsupervised methods in the realms of writer attribution for historical documents and forensic document analysis.

A Mostafa \cite{mostafa2021ocformer} created a custom dataset mimicking historical Arabic document noise and achieved significant progress in layout recognition, segmentation, and character recognition. The model, adept at transcribing handwritten manuscripts and detecting Arabic diacritics, demonstrated impressive results with a Character Error Rate (CER) of 0.0727, Word Error Rate (WER) of 0.0829, and Sentence Error Rate (SER) of 0.10.

M Badry\cite{badry2021quranic} applied machine learning to recognize IoT sensor data, focusing on optical character recognition (OCR) for handwritten Arabic scripts. They proposed two deep learning models—sequence-to-sequence and fully convolutional. Assessing these models on the QTID dataset, the first Arabic dataset with diacritics, they surpassed benchmarks like Tesseract and ABBYY FineReader, achieving state-of-the-art results in character recognition rate, F1-score, precision, and recall. The proposed model achieved a character recognition rate of approximately 97.60\% and 97.05\% for text with and without diacritics, respectively, with an overall recognition rate of 99.48\%. The CNN model exhibited a CRR of approximately 98.90\% and 98.51\% for text with and without diacritics, respectively.

EA El-Sherif\cite{el2007two} presents the introduction of a comprehensive Arabic Handwritten Digits Database (AHDBase), encompassing 60,000 digits for training and 10,000 digits for testing, contributed by 700 individuals with diverse demographic characteristics. Additionally, the authors propose a recognition system tailored for Arabic handwritten digits. The recognition system comprises two stages. In the initial stage, an Artificial Neural Network (ANN) is deployed, utilizing a concise yet potent feature vector for swift classification of non-ambiguous cases. Notably, the first stage incorporates a reject option to channel ambiguous cases to the second stage. The subsequent stage employs a Support Vector Machine (SVM), characterized by a slower yet more powerful nature. This stage utilizes a larger feature vector to effectively classify cases rejected by the first stage, particularly those with greater complexity.

\subsection{Scanned Documents}
Extracting information from scanned documents poses challenges due to their rough layout and low resolution compared to regular documents. Successfully preprocessing the scanned documents is crucial for accurate information extraction, as it plays a significant role in this context. An example of an initiative addressing this challenge is the competition organized by ICDAR Z Huang\cite{huang2019icdar2019}. This competition focuses on extracting information from 1000 scanned receipts and encompasses operations like text identification, arrangement examination, and data retrieval. By participating in such competitions, researchers and practitioners aim to advance the field of information extraction from scanned documents.

\subsection{Quranic Text}
MH Bashir \cite{bashir2023arabic} focused on the Qur'an, a sacred text in the Arabic language, read and followed by nearly two billion Muslims worldwide. As Islam rose to prominence, Arabic became a widespread language across large regions. Devout Muslims turn to the Qur'an daily for guidance. Despite the brevity of the Qur'an itself, an extensive body of supporting work, including commentaries and exegesis, spans tens of thousands of volumes. Recent advancements in computational and natural language processing (NLP) techniques have sparked a renewed interest in these religious texts, particularly among non-specialists. The paper provides a comprehensive survey of Qur'anic NLP efforts, encompassing tools, datasets, and approaches. The scope ranges from automated morphological analysis to the correction of Qur'anic recitation through speech recognition. The authors discuss multiple approaches for various tasks and outline potential future research directions in the field. Trained with diverse recitations and using mel-frequency cepstral coefficients (MFCCs), the model achieves an outstanding 99.89\% recognition rate for 3-second Quranic recitations. This surpasses other techniques in the study, advancing Quranic NLP research for accurate and proficient recitation.

R Malhas\cite{malhas2022arabic} diligently tackled the scarcity of Arabic datasets for machine reading comprehension on the Holy Qur'an by creating QRCD, the pioneering Qur'anic Reading Comprehension Dataset. This dataset, meticulously crafted, contained 1,337 question-passage-answer triplets, incorporating 14\% multi-answer questions to capture real-world complexities. To augment existing models, they introduced CLassical-AraBERT (CL-AraBERT), thoughtfully pre-trained on Classical Arabic data, enriching the Modern Standard Arabic (MSA) resources. Through meticulous cross-lingual transfer learning, they expertly fine-tuned CL-AraBERT using MSA-based MRC datasets and QRCD, leading to the development of the very first MRC system tailored for the Holy Qur'an. Their innovative evaluation metric, Partial Average Precision (pAP), ingeniously accommodated partial matching in both multi-answer and single-answer MSA questions, demonstrating their keen attention to detail and commitment to precise evaluation methodologies.

\section{Optical Character Recognition (OCR) Processes in Arabic}
OCR, or Optical Character Recognition, is a complex process that transforms text characters from printed or handwritten sources into machine-encoded text. This involves several stages: preprocessing, segmentation, recognition, and postprocessing.

In preprocessing, the input image undergoes cleanup and enhancement to optimize recognition quality. Segmentation divides the image into characters, and feature extraction identifies characteristics like shape and size. Recognition classifies characters by comparing them to known ones, and postprocessing corrects errors for refined results. The accuracy of OCR is influenced by factors like image quality, font type, size, and language, leading to varying accuracy levels in different scenarios.

\subsection{Preprocessing}
The accuracy of OCR models can be compromised by formatting issues in images. Challenges like image orientation or color correction problems can notably hinder model performance. To overcome these issues and improve OCR model accuracy during training, image preprocessing techniques are frequently utilized. These techniques are pivotal in optimizing OCR model performance by tackling formatting issues. They include procedures like resizing, grayscale conversion, skew correction, and enhancing image resolution. Implementing these preprocessing techniques improves the quality and suitability of images for OCR training, resulting in enhanced accuracy in text recognition.

AP Tafti\cite{tafti2016ocr} conducted detailed evaluations using popular OCR services: Google Docs OCR, Tesseract, ABBYY FineReader, and Transym. Basic image preprocessing, like grayscale conversion and brightness/contrast adjustments, enhanced recognition accuracy by up to 9\%. Illumination adjustments, emphasizing object sharpness and contour clarity, further improved results.

M Eltay\cite{eltay2022generative} introduced an adaptive data augmentation method utilizing Generative Adversarial Networks (GANs) to address class imbalance in text recognition tasks. Specifically focusing on handwritten Arabic text recognition, the research presented experimental findings from two public datasets. The study demonstrated the effectiveness of GANs trained with this technique in managing class imbalances. 
IB Mustapha\cite{mustapha2022conditional} introduced a method called Conditional Deep Convolutional Generative Adversarial Networks (CDCGAN) designed for generating isolated handwritten Arabic characters. Their experiments, both qualitative and quantitative, demonstrated that CDCGAN effectively produces synthetic handwritten Arabic characters.

\subsection{Segmentation}
Segmentation is a pivotal step in Optical Character Recognition (OCR), involving the intricate division of an image into its constituent parts—lines, words, and characters—to facilitate accurate recognition. Three fundamental segmentation techniques are commonly applied: line segmentation, word segmentation, and character segmentation. Word segmentation, particularly challenging in cursive languages like Arabic, poses specific difficulties due to the absence of clear boundaries between characters. Traditional methods in Arabic OCR rely heavily on rule-based approaches and heuristics that leverage character features. However, the advent of deep learning techniques, including convolutional and recurrent neural networks, has showcased promising advancements in automatic segmentation.

The precision of segmentation is paramount for achieving accurate recognition results. The improvement of Arabic OCR segmentation techniques carries substantial implications for various applications, ranging from document digitization to text-to-speech conversion and language translation. Through the refinement of segmentation methods, the efficiency and effectiveness of Arabic OCR systems can experience notable enhancement, leading to improved performance in diverse language processing tasks.

E Mohamed\cite{mohamed2019arabic} presents Arabic-SOS, a system for pre-processing pre-Modern Arabic text by focusing on segmentation and orthography standardization. The system utilizes the Gradient Boosting algorithm for training a morphological segmenter and a machine learner for text standardization. The results demonstrate high accuracy in segmentation and orthography standardization, outperforming other segmenters on a classical test set.

HA Abdo\cite{abdo2022approach} introduced a novel approach for the analysis of Arabic text documents, a critical process in the development of Optical Character Recognition (OCR) systems. Their approach comprises four key steps: preprocessing, text line segmentation, word segmentation, and character segmentation. The horizontal projection method is employed to detect and extract text lines from preprocessed documents. In the word segmentation step, the computation of space thresholds determines spaces between connected components, facilitating the segmentation of text lines into isolated words. Finally, a thinning method is applied to identify the skeleton of segmented words and analyze geometric characteristics to detect ligatures and characters. The proposed approach was rigorously tested on a set of 115 text images, including those from the KHATT database and images produced by the authors. The experimental results are highly promising, achieving a success rate of 98.6\% for line segmentation, 96\% for word segmentation, and 87.1\% for character segmentation.

S Naz\cite{naz2016segmentation} delved into the realm of Arabic script-based text recognition, an extensively researched field with applications in the educational process for students and educators seeking to comprehend educational content in Arabic script. Despite its long-standing prominence, the challenging nature of Arabic script recognition has been a focal point for researchers in both industry and academia. However, these endeavors have yet to yield satisfactory results. The authors particularly focused on the intricacies of segmenting Urdu script in the Nasta’liq writing style, a formidable task given its complexity compared to the Naskh writing style. Emphasizing the critical role of segmentation for achieving high accuracy, the study highlighted character segmentation as a pivotal phase in the Optical Character Recognition (OCR) process. The authors underscored the importance of segmentation by noting higher recognition rates for isolated characters compared to results for words or connected characters. The current study specifically investigated recent advancements in character segmentation and the challenges associated with segmenting languages based on the Arabic script.

\subsubsection{Line segmentation}
In the Optical Character Recognition (OCR) process, line segmentation is a crucial step that involves dividing the skew-corrected image into separate lines of text. This is essential because it allows the OCR system to process each line independently, leading to improved recognition of text within the image. Line segmentation can be achieved through various techniques, such as connected component analysis, project profile analysis, and machine learning approaches. By segmenting the text into lines, the OCR system can enhance the accuracy of character recognition.

A Qaroush\cite{qaroush2022learning}  addressed the challenging task of extracting text lines from document images, a crucial step in optical character recognition and an ongoing issue in document analysis. They specifically tackled the complexities arising from diverse font variations, diacritics, and overlapping or touching text lines, which pose challenges for algorithms designed for machine-printed text. The authors introduced a simple yet robust two-stage algorithm for text-line extraction in printed Arabic text. The method efficiently extracts text lines, even in small font sizes, leveraging baselines, projection profiles, and a top-down divide and conquer technique. Notably, the proposed algorithm demonstrated effectiveness in handling non-uniform inter-line spacing and the overlapping problem. Through a series of experiments on a collected dataset, the authors compared the proposed method with two baseline approaches. The results showed that the proposed algorithm outperformed the baselines, achieving an average error rate of 3\% for Arabic text without diacritics and 11\% for Arabic text with diacritics. Additionally, the experiments highlighted the computational efficiency of the proposed algorithm, with an average running time of 0.087 seconds per document image.

\subsubsection{Word segmentation}

After completing line segmentation, the next phase involves word segmentation, which includes breaking down the line of text into individual words. Diverse methods are utilized for word segmentation. The Spacing method utilizes spaces between words to delineate and segment the text. Conversely, the dictionary-based method employs a word dictionary for comparison, identifying word boundaries by matching against the text. Character-based methods rely on recognized character patterns, like word breaks and punctuation, to execute word segmentation.

M Boualam\cite{boualam2022arabic} developed an end-to-end system utilizing a deep Convolutional Recurrent Neural Network (CNN/RNN). Their system was trained on the IFN/ENIT extended database to enhance its performance.

M Elkhayati\cite{elkhayati2022segmentation}  proposes a method to segment handwritten Arabic words into graphemes using a directed Convolutional Neural Network (CNN) and Mathematical Morphology Operations (MMO). Arabic's cursive script requires link removal for accurate segmentation. The study addresses challenges like diacritics and over-traces, introducing solutions: overcoming over-traces, robust diacritics extraction, and using a Partial Dilation (PD)-Global Erosion (GE) technique for segmentation. PD enhances vital areas, while GE removes inter-grapheme links, ensuring accurate segmentation despite information loss. The method utilizes a directed CNN for robustness.

MA Sabri\cite{sabri2023robust} developed a method to automatically separate text from graphics in printed Arabic historical documents, a vital step for digitization. They introduced a fast and efficient technique using hybrid modeling involving graphs and structural analysis. The method employed the RLSA smoothing algorithm for segmentation and utilized the DTW (Dynamic Time Warping) algorithm for matching word features.

\subsubsection{Character segmentation}
Character segmentation is a method used to divide an image of a single word into distinct letters and characters. Its applicability depends on the specific requirements of the OCR system in use. If the text comprises separate and well-defined letters within a word, character-level segmentation may be unnecessary as the previous segmentation step can adequately handle letter and character separation using a thresholding approach. However, in cases where the text exhibits cursive handwriting or connected letter forms, character-level segmentation becomes essential.

NH Khan\cite{khan2018urdu} conducted a comprehensive review and survey of the major studies in Urdu optical character recognition (OCR). The paper begins by introducing OCR technology and providing a historical overview of OCR systems, drawing comparisons between English, Arabic, and Urdu systems. Extensive background and literature are presented for the Urdu script, covering its history, OCR categories, and phases. The study reports on the latest advancements in different OCR phases, including image acquisition, pre-processing, segmentation, feature extraction, classification/recognition, and post-processing for Urdu OCR systems. Emphasis is placed on segmentation, discussing both analytical and holistic approaches for Urdu text. The feature extraction section includes a comparison between feature learning and feature engineering approaches, covering both deep learning and traditional machine learning methods. The paper also touches on Urdu numeral recognition systems. In conclusion, the research paper identifies open problems in the field and suggests future directions for Urdu OCR research.

M Tayyab\cite{tayyab2022recognition} addresses the significant area of news ticker recognition, recognizing its importance for applications like information analysis, opinion mining, and language translation, particularly for media regulatory authorities. The focus of the study is on developing an automated Arabic and Urdu news ticker recognition system, encompassing ticker segmentation and text recognition to extract textual data for various online services. The research delves into character-wise explicit segmentation and syntactical models, considering Kufi and Nastaleeq fonts. Various network models are employed to learn deep representations, homogenizing classes regardless of inter-symbol correlations and linguistic taxonomy. The proposed learning model emphasizes fairness by maximizing balance among sensitive features of characters in a unified manner. The efficiency of the model is demonstrated through experiments using customized news tickers datasets with accurate character-level and component-level labeling. Additionally, the method is evaluated on the challenging Urdu Printed Text Images (UPTI) dataset, which only provides ligature-based annotations. The proposed method achieves a notable accuracy of 98.36\%, surpassing the current state-of-the-art method. Ablation investigations highlight that the technique notably improves the performance of character classes with low symbol frequencies.

OA Boraik\cite{boraik2022characters} proposed a hybrid approach for accurate segmentation of interconnected characters. Their method involved stages like removing extra shapes, detecting individual words using morphological operations, and employing computational analysis for touching character segmentation. Tests were conducted on various datasets, including KHATT and IFN/ENIT, showcasing the method's effectiveness.

\subsection{Recognition}
Recognition, also known as classification in earlier studies, is a pivotal stage in Optical Character Recognition (OCR) that involves identifying and assigning specific characters or character groups to an input image. After the preprocessing and segmentation phases, the OCR system compares the extracted features of each character or character group with a set of predefined templates or models. During the training phase, the system constructs these templates using a substantial dataset of sample images to teach the system accurate character recognition.

The classification or recognition process incorporates various algorithms, such as template matching, neural networks, and support vector machines. Template matching entails comparing the extracted features of each character with predefined templates and selecting the template that closely corresponds to the recognized character. In contrast, neural networks and support vector machines use machine learning algorithms to learn and classify patterns in the data, often achieving higher accuracy compared to template matching.

T Milo\cite{milo2019new} proposed an innovative Optical Character Recognition (OCR) strategy for Arabic, addressing current limitations. They introduced the concept of archigraphemes, considering Arabic script as an allographic rendering. A minimum unit, the letter block, formed the basis for script grammar and OCR. An algorithm reduced Unicode text to archigraphemes, enabling the synthesis of realistic images. The approach laid the groundwork for an OCR system, emphasizing controlled conditions and extended shaping. The study highlighted the need for further training, linguistic parsing of archigraphemes, and outlined steps towards static OCR technology, with future plans to make the OCR dynamic using AI software.

\subsubsection{Character Recognition}
Table \ref{tab:character recognition Methods} lists all the Character Recognition strategies.  It also discusses various deep learning-based methods that have been used in these methods.

L Bouchakour\cite{bouchakour2021printed} explored Optical Character Recognition (OCR) for the Arabic language, a complex task due to the intricacies of Arabic script. Their OCR system comprises segmentation, feature extraction, and recognition stages. They proposed a novel method for recognizing printed Arabic characters, utilizing combined feature extraction techniques including densities of black pixels, Hu and Gabor features' invariant moments, and a Convolutional Neural Network (CNN) classifier. 

N Altwaijry\cite{altwaijry2021arabic} introduced a novel dataset named Hijja, comprising 47,434 characters written by 591 children aged 7–12. Alongside the dataset, they proposed an automatic handwriting recognition model based on convolutional neural networks (CNN). The model was trained on both the new Hijja dataset and the Arabic Handwritten Character Dataset (AHCD), with accuracies of 97\% and 88\% on the AHCD dataset and Hijja dataset, respectively

BH Nayef\cite{nayef2022optimized} proposed an optimized leaky rectified linear unit (OLReLU) to improve the handling of imbalanced positive and negative vectors, a common issue in handwritten character recognition. The proposed OLReLU was integrated into a CNN architecture with a batch normalization layer to enhance overall performance. Evaluation was conducted on four datasets, including the Arabic Handwritten Characters Dataset (AHCD), self-collected data, Modified National Institute of Standards and Technology (MNIST), and AlexU Isolated Alphabet (AIA9K). The proposed method demonstrated significant improvements in terms of accuracy, precision, and recall when compared to state-of-the-art methods, achieving remarkable results on AHCD (99\%), self-collected data (95.4\%), HIJJA dataset (90\%), and Digit MNIST (99\%). 

YM Alwaqfi\cite{alwaqfi2022generative} addressed the complexity of Arabic character recognition by exploring generative adversarial networks (GANs), a compelling algorithm in neural networks. They proposed utilizing the sigmoid cross-entropy loss function for recognizing Arabic handwritten characters through multi-class GANs training algorithms. The evaluation, conducted on a dataset of 16,800 Arabic handwritten characters, demonstrated the effectiveness of the proposed approach, achieving an impressive accuracy of 99.7\%. This research contributes to advancing Arabic character recognition and highlights the potential of multi-class GANs in enhancing recognition accuracy.

M Alheraki\cite{alheraki2023handwritten}  introduced a CNN model designed to recognize children's handwriting. The model achieved a remarkable accuracy rate of 91\% on the Hijja dataset, comprising Arabic characters written by children, and 97\% on the Arabic Handwritten Character Dataset.

EF Bilgin Tasdemir \cite{bilgin2023printed} introduces a deep learning (DL)-based system for character recognition of printed Ottoman script. The study begins by creating a synthetic text image dataset sourced from a text corpus, which is then augmented using various image processing techniques. A hybrid architecture combining convolutional neural network (CNN) and bidirectional long short-term memory (BiLSTM) models is developed for the character recognition task. The recognizer is trained using both the original dataset and the augmented dataset. To further enhance the system's performance on real image data, a transfer learning procedure is applied. This procedure enables the adaptation of the DL-based character recognition system to real-world image data, resulting in improved accuracy and robustness. The presented work contributes to the advancement of character recognition technology specifically tailored for the printed Ottoman script.

A Bin Durayhim\cite{bin2023towards} aimed to develop models for recognizing children's Arabic handwriting. They introduced two deep learning-based models: a Convolutional Neural Network (CNN) and a pre-trained CNN (VGG-16). These models were trained using the Hijja dataset, which comprises recent samples of Arabic children's handwriting collected in Saudi Arabia. Additionally, they conducted training and testing using the Arabic Handwritten Character Dataset (AHCD) for evaluation. Comparative analysis with similar models from existing literature demonstrated that their proposed CNN model outperformed both the pre-trained CNN (VGG-16) and other models considered. Furthermore, the authors created a prototype called Mutqin, designed to assist children in practicing Arabic handwriting. The prototype underwent evaluation by its intended users, and the study reports the results of this evaluation.

\begin{table}[h!]
\caption{A comparison of the benefits and drawbacks of several deep learning-based Character Recognition methods}
\begin{adjustbox}{width=1\textwidth}
%\resizebox{!}{0.3\textheight}{
\begin{tabular}{|c|c|c|c|} 
    \hline
      Literature & Method & Benefits & Drawbacks  \\ \hline
      
      L Bouchakour\cite{bouchakour2021printed}         & \makecell{Combined Hu and Gabor\\ features + CNN classifier}     &     \makecell{ The method explores the use of combined features,\\ resulting in improved recognition rates compared to\\ using a single feature.}     & \makecell{1) Computationally intensive. 2) Require la-\\rge amounts of labeled data for effective\\ training.}\\ \hline

      N Altwaijry\cite{altwaijry2021arabic}  & CNN    &     \makecell{ 1) Introduce a benchmark dataset Hijja. 2) Authors\\ proposed a deep learning model to recognize hand-\\written letters}     & \makecell{They didn't examine successful machine le-\\arning models like MLP, SVM, and auto-\\encoders on this dataset.}\\ \hline

      BH Nayef\cite{nayef2022optimized} &  \makecell{CNN + optimized leaky re-\\ctified linear unit (OLReLU)}   &    \makecell{The model maintains a balance between positive and \\negative feature maps while training the CNN.} & \makecell{OLReLU requires more time compared to \\using ReLU.}\\ \hline
      
      YM Alwaqfi\cite{alwaqfi2022generative}  &  \makecell{Generative Adversarial Netw-\\orks (GANs) + Deep CNN}   &    \makecell{1) GANs generate diverse images for dataset augment-\\ation, enhancing model robustness. 2) Stability during \\training is achieved using activation functions and dr-\\opout layers.} & \makecell{1) Model performance heavily relies on the \\quality of the generated dataset. 2) Add-\\ition of dropout layers may decrease model \\accuracy, especially when used with batch\\ normalization.}\\ \hline

      M Alheraki\cite{alheraki2023handwritten}         & \makecell{Convolutional Neural Netw-\\ork (CNN)}     &     \makecell{1) The adoption of both single and multi-model app-\\roaches allows flexibility in training strategies, potentia-\\lly improving recognition accuracy. 2) The LeakyReLU \\activation function, with a slope of 0.3, addresses the \\vanishing gradient problem, leading to improved results \\compared to the baseline.}     & \makecell{1) The model's reliance on accurate stroke\\ count information may make it sensitive\\ to variations in the quality of the gener-\\ated dataset. 2) The provided informat-\\ion lacks specific details on the number \\of neurons in certain layers, making it ch-\\allenging to assess the complexity of the \\architecture comprehensively. }\\ \hline

      EF Bilgin Tasdemir \cite{bilgin2023printed}        & \makecell{CNN-BLSTM-CTC(Conne-\\ctionist Temporal Classif-\\ication)}     &     \makecell{ 1) The hybrid architecture combines CNN-BLSTM-CTC\\ has shown success in sequence labeling tasks. 2) BLS-\\TM networks are effective in capturing temporal depen-\\dencies in sequence labeling tasks, contributing to the\\ model's ability to recognize the structure of Ottoman \\text. }     & \makecell{1) The synthetic dataset's size is constra-\\ined by the source text corpus, limiting\\ the model's capacity for diverse learning.\\ 2) The system's effectiveness heavily dep-\\ends on the quality and quantity of the \\training data}\\ \hline

      A Bin Durayhim\cite{bin2023towards}         & \makecell{CNN , VGG-16}     &     \makecell{ 1) Attains superior accuracy on standardized datasets.\\ 2)Streamlines training processes, thereby optimizing\\ model efficiency.}     & \makecell{1) Computationally intensive. 2) Addition-\\al post-processing processes are necessary}\\ \hline

\end{tabular}%}
\end{adjustbox}
\label{tab:character recognition Methods}
\end{table}

\subsubsection{Word Recognition}
Table \ref{tab:word recognition Methods} lists all the Character Recognition strategies.  It also discusses various deep learning-based methods that have been used in these methods.

K Hamad \cite{hamad2016detailed} conducted a comprehensive investigation into Optical Character Recognition (OCR), recognizing its significance in various fields for storing and retrieving information from printed or handwritten documents. They emphasized the challenges associated with OCR, including font characteristics and image quality. The paper provided a detailed overview of OCR stages, covering pre-processing, segmentation, normalization, feature extraction, classification, and post-processing. Additionally, the authors explored the historical development of OCR, its main applications, and recent advancements. This comprehensive review aims to contribute a thorough understanding of the challenges and advancements in OCR technology, addressing the complexities of extracting and processing text from diverse document formats.

M Awni\cite{awni2019deep} employs model averaging as an ensemble learning method, training three Residual Networks (ResNet18) models. The core concept involves combining diverse classifiers to compensate for individual mistakes, ultimately improving the ensemble's prediction accuracy. The ResNet18 architecture is utilized for its balance between depth and performance, with modifications to align the output nodes with the distinct words present in the IFN/ENIT database. The optimization techniques employed include gradient descent, Root Mean Square Propagation (RMSProp), and Adaptive Momentum Estimation (Adam). To efficiently determine the learning rate, the authors employ a Learning Rate Range Finder test, crucial for training convolutional neural networks. Finally, the ensemble model for offline Arabic handwritten word recognition incorporates variations in optimization techniques for each member, employing model averaging for the final prediction. The methodology is validated through experiments on the IFN/ENIT database, demonstrating the effectiveness of the proposed ensemble approach.

SK Jemni\cite{jemni2019out} addressed the issue of out-of-vocabulary (OOV) words in Arabic handwriting recognition systems, which typically operate with a fixed vocabulary. They proposed a two-step approach to tackle this problem. In the first step, various sub-word units were employed to identify potential OOVs. In the recovery stage, a dynamic dictionary was created to augment the initial static word lexicon, accommodating the detected OOVs. The recovery process involved selecting the best word candidates from an external resource. The experiments conducted on the KHATT and AHTID/MW databases demonstrated that sub-word modeling contributed to better detection, and the use of a dynamic dictionary significantly enhanced recognition performance compared to one-step approaches based on a large static dictionary or the combination of different sub-word units. The proposed approach achieved state-of-the-art results on the KHATT dataset.

M Eltay\cite{eltay2020exploring} introduced an adaptive data augmentation algorithm assigning weights to words based on class probabilities. The RNN-LSTM architecture, configured with LSTM layers and employing the word beam search algorithm, demonstrated state-of-the-art results. They adapted the AlexNet for the task, replacing the last layer and employing transfer learning. Additionally, the authors proposed a method for out-of-vocabulary (OOV) word detection and recovery, achieving improved recognition performance compared to existing approaches.

H Butt\cite{butt2021attention} proposed a CNN-RNN model with an attention mechanism for Arabic image text recognition. The model utilizes a CNN to generate feature sequences from input images, followed by a bidirectional RNN to arrange these sequences. To enhance text segmentation, a bidirectional RNN with an attention mechanism is employed to produce the final output. This attention mechanism enables the model to selectively focus on relevant information within the feature sequences. The end-to-end training is implemented through a standard backpropagation algorithm, showcasing the effectiveness of the proposed approach in Arabic image text recognition.

S Bergamaschi\cite{bergamaschi2022novel} addresses the imperative need for managing multicultural heritages in a globalized context. The DigitalMaktaba project, a collaborative effort among computer scientists, historians, librarians, engineers, and linguists, aims to establish efficient procedures for cataloging archival heritage in non-Latin alphabets. The paper discusses the ongoing development of a novel workflow and tool for text sensing, focusing on automatic knowledge extraction and cataloging of documents in languages like Arabic, Persian, and Azerbaijani. The prototype utilizes advanced OCR, text processing, and information extraction techniques to ensure accurate text extraction and rich metadata content, surpassing current limitations. 

N Alzrrog\cite{alzrrog2022deep}  addressed the limited progress in research on automatic Arabic handwriting word recognition using deep learning neural networks. They highlighted the absence of a general and reliable Arabic Handwritten words database, which hinders the advancement of related research. To fill this gap, the authors introduced a new Deep Convolutional Neural Network (DCNN) algorithm applied to a novel dataset called Arabic Handwritten Weekdays Dataset (AHWD). This dataset, consisting of 21,357 words distributed among seven classes, was meticulously prepared by 1,000 individuals. The proposed DCNN model was trained on this balanced dataset using various structures and enhanced with techniques like dropout, image regularization, and proper learning rates to prevent overfitting. The model demonstrated promising performance, achieving an accuracy rate of 99.39\% with an error rate of 4.61\% on the AHWD dataset and an accuracy rate of 99.71\% with an error rate of 1.71\% on the IFN/ENIT dataset. The authors conducted a blind test on the hidden test set and assessed performance using a confusion matrix and learning curves, affirming the model's effectiveness in Arabic handwriting word recognition.

GJ Salman \cite{salman2023proposed} tackled the challenging task of recognizing Arabic words and texts, especially when presented in varying sizes and fonts. They developed an intelligent system that involved the creation of a comprehensive dataset comprising 1,000 words, each written in 24 different ways using various Arabic fonts. Leveraging image processing methods, the system identified and deduced words from images. The core of the system relied on a deep learning approach, specifically a Convolutional Neural Network (CNN) algorithm. This CNN algorithm was trained to extract features from truncated words and retrieve text words that closely resembled the ones that were cut. In experimental evaluations, the system demonstrated remarkable performance, achieving a 99\% accuracy in word detection and a 96\% accuracy in word recognition.

I Ouali \cite{ouali2023augmented}  present a novel system designed to recognize and identify Arabic Handwritten Texts with Diacritics (AHTD) through the utilization of deep learning, specifically the convolutional neural network. The system is trained, tested, and validated using our developed Arabic Handwritten Texts with a Diacritical Dataset (AHT2D). Following the recognition process, the identified text undergoes enhancement through augmented reality (AR) technology, resulting in a visually enhanced 2D image representation. Furthermore, the authors leverage AR technology to convert the recognized text into an audio output, catering to the needs of visually impaired users. By providing both voice and visual outputs, our system offers an inclusive solution to facilitate effective communication and accessibility for visually impaired individuals.

S Hamida\cite{hamida2023cursive} introduced a novel image processing approach that combines three image descriptors to extract features from handwritten Arabic text. Using a subset of 100 classes from the IFN/ENIT handwritten Arabic database, they applied preprocessing techniques and trained separate k-nearest neighbor (k-NN) models for each feature descriptor. These models were used to classify Arabic handwritten images, and their performance was evaluated using common metrics. Impressively, the final model achieved an exceptional recognition rate of up to 99.88\%

S Malakar\cite{malakar2023handwritten} focused on handwritten word recognition (HWR), a persistent challenge in the field. The study chose the holistic approach for its effectiveness with limited lexicons. It noted the absence of consideration for inter-segment similarity in existing methods, which could offer valuable insights. To address this gap, the authors introduced Hausdorff and Fréchet distances to quantify similarity among different word segments. They also used shape-based descriptors and combined outputs from six classifiers using majority voting. Evaluation on standard databases (IAM and IFN/ENIT) yielded promising results compared to state-of-the-art HWR methods.

YM Alwaqfi\cite{alwaqfi2023novel} addressed challenges in recognizing printed Arabic words due to the language's complexity and limited available datasets. They proposed a hybrid model combining a deep convolutional neural network (DCNN) as a classifier and a generative adversarial network (GAN) for data augmentation. This hybrid model significantly improved accuracy and generalization ability, achieving a remarkable accuracy score of 99.76\% on the Arabic printed text image dataset (APTI) compared to 94.81\% with the DCNN alone.

\begin{table}[h!]
\caption{A comparison of the benefits and drawbacks of several deep learning-based Word Recognition methods}
\begin{adjustbox}{width=1\textwidth}
%\resizebox{!}{0.3\textheight}{
\begin{tabular}{|c|c|c|c|} 
    \hline
      Literature & Method & Benefits & Drawbacks  \\ \hline

      M Awni\cite{awni2019deep} & \makecell{Ensemble of Residual Net-\\works (ResNet18)}     &     \makecell{More robust to overfitting and noise in the data}     & \makecell{Need more post-processing processes}\\ \hline

      SK Jemni\cite{jemni2019out} & \makecell{CNN-MDLSTM with \\Dynamic lexicons (DWLD)}      &     \makecell{More robust to overfitting and noise in the data}     & \makecell{Need more post-processing processes}\\ \hline

      M Eltay\cite{eltay2020exploring} & \makecell{Bidirectional Long Short-\\Term Memory(BLSTM)-Co-\\nnectionist Temporal Class-\\ification(CTC)-Word Beam \\Search(WBS)}     &     \makecell{Solve effectively the data imbalance problem}     & \makecell{The method used leads to increase in \\the training time of the network}\\ \hline
      
      H Butt\cite{butt2021attention}      & \makecell{CNN-RNN with Atten-\\tion for Arabic Image Text \\Recognition}     &     \makecell{1) Bidirectional RNNs handle sequential data, capturing \\contextual dependencies in Arabic text.}     & \makecell{Need higher computational requirements.}\\ \hline

      S Bergamaschi\cite{bergamaschi2022novel}     & \makecell{OCR (GoogleDocs, \\EasyOCR, Tesseract)}     &     \makecell{1) The tool speeds up document cataloguing, reducing\\ manual effort significantly, especially for simpler docu-\\ments 2) Automatic suggestions enhance accuracy, m-\\inimizing errors during data insertion and ensuring co-\\nsistent catalogued information.}     & \makecell{1) The system's performance relies on i-\\mage quality, potentially limiting effect-\\iveness with suboptimal images. 2) Lim-\\ited development in Arabic-script text s-\\ensing poses challenges for full automa-\\tion in this context. 3) Large data loads \\are required for machine learning featur-\\es, demanding a substantial initial man-\\ual workload.}\\ \hline

      N Alzrrog\cite{alzrrog2022deep} & \makecell{Deep Convolutional Neural\\ Network (DCNN)}      & \makecell{ The model demonstrates a high accuracy rate. } &  \makecell{1) Making a dataset private means we \\cannot experience their results.}  \\ \hline

      GJ Salman \cite{salman2023proposed}     & \makecell{Multi-Font Arabic Word\\ Recognition CNN}     &     \makecell{Inclusion of a dataset with 1,000 words written in 24 \\different ways using various Arabic fonts contributes \\to comprehensive training.}     & \makecell{The need for a diverse dataset with mult-\\iple font variations necessitates extensive\\ data collection and preprocessing efforts.}\\ \hline

      I Ouali \cite{ouali2023augmented}     & \makecell{AHTD based on CNN and \\Augmented Reality (AR)}     &     \makecell{Integrates with smart glasses through a mobile appli-\\cation, allowing users to capture, extract, and listen to\\ text content, enhancing the overall user experience.}     & \makecell{1) Limited discussion on AR engine; m-\\ore details on Vuforia's suitability and al-\\ternatives would improve comprehension.\\ 2) The system assumes user eye move-\\ment for text detection and recognition, \\potentially limiting applicability to all v-\\isually impaired users.}\\ \hline

      S Hamida\cite{hamida2023cursive}     & \makecell{k-nearest neighbor \\algorithm (k-NN)}     &     \makecell{Integrates HOG, GF and LBP for precise feature extr-\\action, enhancing Arabic handwritten text recognition.}     & \makecell{Involves intricate preprocessing steps}\\ \hline
      
      S Malakar\cite{malakar2023handwritten}     & \makecell{ Hausdorff and Fréchet dist-\\ances for inter-segment si-\\milarity features in word r-\\ecognition.}     &     \makecell{1) Introduces a holistic word recognition method using \\inter-segment similarity, offering a unique perspective. \\2) Leverages diverse features like inter-segment simil-\\arity, shape-based, and contour-based for improved\\ recognition accuracy.}     & \makecell{1) Additional post-processing processes \\are necessary. 2) Relies on experimental\\ threshold values, making it sensitive to\\ parameter tuning }\\ \hline

      YM Alwaqfi\cite{alwaqfi2023novel}     & \makecell{Hybrid GAN-based Model\\ + DCNN}     &     \makecell{1) Achieves outstanding performance. 2) The strate-\\gy based on GAN compels the network to extract si-\\milar characteristics.}     & \makecell{1) No comparison with other methods.\\ 2) The model relying on a generator is \\vulnerable when dealing with images \\of varying layouts.}\\ \hline

\end{tabular}%}
\end{adjustbox}
\label{tab:word recognition Methods}
\end{table}

%%%%%%%%%%%%%%%%%%%%%%%%%%%%%%%%%%%%%%%%%%%%%%%%%%%%%%%%%%%%%%%%%%%%%%%%%%%%%%%%%%%%%%%%%%%

\subsubsection{Digits Recognition}
Table \ref{tab:Digits recognition Methods} lists all the Character Recognition strategies.  It also discusses various deep learning-based methods that have been used in these methods.

E Al-wajih\cite{al2020improving} addressed the challenge of Arabic handwritten digit recognition, leveraging sliding windows for enhanced classification accuracy. Employing Random Forests (RF)and Support Vector Machine (SVM) classifiers, they introduced four feature extraction techniques—Mean-based, Gray-Level Co-occurrence Matrix (GLCM), Moment-based, and Edge Direction Histogram (EDH). Through meticulous evaluation, the study demonstrated the efficacy of sliding windows, achieving notable recognition rates, such as 98\% for Mean-based and Moment-based with RF, and 98.33\% and 99.13\% for GLCM and EDH with linear-kernel SVM. Using a modified AHDBase dataset, the proposed model, incorporating various sliding window sizes, offers insights into optimizing feature extraction and classification. Comparative analysis against cutting-edge approaches underscores the significance of this approach in advancing Arabic handwritten digit recognition, providing a foundation for future research in optimal methodology combinations.

YS Can\cite{can2020automatic} focused on the digitization and recognition of Arabic numerals from the initial series of population registers of the Ottoman Empire in the mid-nineteenth century. To isolate numerals written in red, a color filter was applied, leveraging the distinctive structure of the registers. Initially, a convolutional neural network (CNN)-based segmentation method was employed for numeral spotting. Subsequently, the authors curated a local Arabic handwritten digit dataset from the identified numerals and tested a Deep Transfer Learning method on this dataset for digit recognition. 

In the research of F Haghighi\cite{haghighi2021stacking} a novel model for recognizing handwritten digits is introduced to address the challenge posed by the distinctive writing styles of individuals, especially in languages like Persian/Arabic. The proposed model is a stacking ensemble classifier, combining Convolutional Neural Network (CNN) and Bidirectional Long-Short Term Memory (BLSTM) techniques. An innovative aspect of the model involves using the probability vector of the images' class as input for the meta-classifier layer, leveraging BLSTM's ability to understand arrays and vectors. By doing so, the model aims to enhance the accuracy of the deep learning model, particularly in capturing the structural similarities of certain Persian/Arabic digits.

RS Alkhawaldeh\cite{alkhawaldeh2022ensemble} proposed an Ensemble Deep Transfer Learning (EDTL) model, combining two pre-trained transfer learning models, to effectively detect and recognize these digits. The EDTL model demonstrated exceptional performance, achieving up to 99.83\% accuracy, surpassing baseline methods, including deep transfer learning models and ensemble deep transfer learning models. The proposed architecture included parallel and sequence blocks with convolution and pooling operations, incorporating ResNet-50 and MobileNetV2 models. This architecture addressed the vanishing gradient issue through skip connections and depth-wise separable convolutions. The EDTL model efficiently extracted relevant features from noisy handwritten digits. The classification phase involved merging these features into a fully-connected Artificial Neural Network (ANN) for accurate recognition, achieving state-of-the-art performance on popular datasets.

S Ali\cite{ali2023recognition} addresses the challenges in recognizing Persian/Arabic handwritten digits, a task crucial for applications like office automation and document processing. They propose a modified Deep Convolutional Neural Network (DCNN) architecture, incorporating three convolutional layers with features such as batch normalization, pooling, fully connected layers, and dropout regularization to enhance generalization and prevent overfitting. The study utilizes the HODA database, applying preprocessing steps like image smoothing and resizing. Various optimization algorithms, including stochastic gradient descent and Adam, are explored to optimize the DCNN. The research investigates the impact of different epochs on Optical Character Recognition (OCR) performance and determines optimal learning parameters. 

\begin{table}[h!]
\caption{A comparison of the benefits and drawbacks of several deep learning-based Digits Recognition methods}
\begin{adjustbox}{width=1\textwidth}
%\resizebox{!}{0.3\textheight}{
\begin{tabular}{|c|c|c|c|} 
    \hline
      Literature & Method & Benefits & Drawbacks  \\ \hline

      E Al-wajih\cite{al2020improving} & \makecell{sliding windows + rando-\\m forests (RF) + support\\ vector machine (SVM) }   &  \makecell{1) Enhace classification accuracy for Arabic digit\\ images. 2) Comprehensive comparison with recent \\state-of-the-art approaches validates performance.}  & \makecell{Experiments are constrained in image \\sizes and sliding window sizes.} \\ \hline

      YS Can\cite{can2020automatic} & \makecell{Deep Transfer Learning \\(DTL) + CNN}  &  \makecell{The study implemented an automatic Arabic num-\\eral spotting system on historical Ottoman Emp-\\ire population registers with notable accuracy.}  & \makecell{limited applicability of AHDBase, yielding \\lower accuracy (72\% with CNN + MLP) \\than CNN alone for recognizing digits in\\ local datasets.}  \\ \hline

      F Haghighi\cite{haghighi2021stacking}    & \makecell{CNN + Bidirectional\\ Long-Short Term Mem-\\ory (BiLSTM)}     &     \makecell{1) provides a holistic solution to the challenges \\posed by diverse writing styles. 2) The model\\ utilizes a probability vector of images' class as\\ input for the meta-classifier layer, enhancing \\feature extraction capabilities, especially in c-\\apturing structural similarities.}     & \makecell{1) To provide equivalent results, many p-\\rocessing stages are necessary. 2) The \\research acknowledges a weakness in d-\\ata processing speed, primarily  attribut-\\ed to the size of input data}\\ \hline

      RS Alkhawaldeh\cite{alkhawaldeh2022ensemble}     & \makecell{Ensemble Deep Transf-\\er Learning (EDTL)}     &     \makecell{1)  Reducing time and cost complexities during\\ training. 2)Effectively extracted relevant features \\from noisy handwritten digits. 3) Combined the \\strengths of two pre-trained models, resulting\\ in enhanced model performance}     & \makecell{1) The Need for effective tuning of hyper-\\parameters to achieve optimal results.\\ 2) The combined size of ResNet-50 and\\ MobileNetV2 may result in a larger model\\ size, impacting deployment in resource-\\constrained environments.}\\ \hline

      S Ali\cite{ali2023recognition}     & \makecell{Modified Deep Convol-\\utional Neural Network \\(DCNN)}     &     \makecell{1) Effective handling of sparse gradients.  2) A \\streamlined and effective approach for Arabic \\digit recognition }     & \makecell{Larger feature sizes increase training t-\\ime; the proposed approach, tested on \\40x40-pixel images, may need validat-\\ion for alternative sizes.}\\ \hline

\end{tabular}%}
\end{adjustbox}
\label{tab:Digits recognition Methods}
\end{table}

%%%%%%%%%%%%%%%%%%%%%%%%%%%%%%%%%%%%%%%%%%%%%%%%%%%%%%%%%%%%%%%%%%%%%%%%%%%

\subsubsection{Multifaceted Recognition Approaches}
Table \ref{tab:Multifacet recognition Methods} lists all the Character Recognition strategies.  It also discusses various deep learning-based methods that have been used in these methods.

N Alrobah \cite{alrobah2022arabic} conducted a comprehensive survey of research projects and experiments that aim to develop machines capable of automatically recognizing handwritten characters, with a specific focus on the unique challenges posed by Arabic script. Unlike studies primarily conducted in Latin scripts, recognizing handwritten Arabic characters is inherently complex due to the nature of Arabic words. The paper particularly delves into recent advancements in the application of deep learning approaches to Arabic character recognition. The main objective is to categorize, analyze, and present a systematic survey of state-of-the-art methods, with a special emphasis on deep learning for feature extraction in offline text recognition. The analysis critically evaluates existing literature, identifies challenges and problem areas, proposes a new classification of the literature, and engages in a thorough discussion of the issues and challenges associated with recognizing Arabic scripts.

R Ahmed\cite{ahmed2021novel} propose a novel context-aware model based on deep neural networks, specifically a supervised Convolutional Neural Network (CNN). This CNN model is designed to contextually extract optimal features, incorporating batch normalization and dropout regularization parameters to prevent overfitting and enhance generalization. The architecture utilizes deep stacked-convolutional layers, forming the proposed Deep CNN (DCNN). The model undergoes comprehensive evaluations across six benchmark databases, showcasing its superior classification accuracy compared to conventional OAHR approaches. Transfer learning (TL) is employed for feature extraction, demonstrating the model's superiority over state-of-the-art pre-trained VGGNet-19 and MobileNet models.

A Mortadi\cite{mortadi2023alnasikh} proposed using the TrOCR architecture, a deep learning model based on transformers, known for its superior performance in text recognition. The authors customized the TrOCR model's encoder and decoder specifically for Arabic script, considering its unique features. Through rigorous experimentation, their approach demonstrated outstanding accuracy.TrOCR incorporates pre-trained computer vision (ViT-style) and natural language processing (BERT-style) models for initializing the encoder and decoder. The encoder processes text line images, partitioning them into patches, which are then flattened, projected into a D-dimensional vector, and combined with positional embedding. The decoder, equipped with multi-head self-attention and feed-forward neural network (FFN) layers, features an encoder-decoder multi-head attention layer. The proposed model, based on TrOCR, modifies the encoder to employ a Vit model and integrates the BERT-Small model's self-attention and FFN layers for the decoder. When applied to recognizing text from scanned Arabic documents at the word level, the approach achieved remarkable precision. The character error rate (CER) was 0.8, indicating precise character recognition, while the word error rate (WER) was 2.3, signifying accurate transcription of complete words. These results highlight the effectiveness of their tailored method in capturing the intricacies of Arabic script during OCR tasks.

HM Al-Barhamtoshy\cite{al2023arabic} introduces a novel framework leveraging the Fast Gradient Sign Method (FGSM) in Keras and TensorFlow for enhanced Arabic manuscript recognition in OCRing systems. The framework focuses on improving OCRing accuracy through deep learning in a multilingual context, optimizing image enhancement, alignment, and layout analysis. RoI detection, performed using a custom-trained deep learning model with bounding box regression, extends the Page Segmentation Method (PSM). The proposed framework, featuring a CNN architecture for adversarial training and FGSM implementation, achieves significant OCRing accuracy improvements, especially in language identification, document category, and diverse RoI types. The model exhibits resilience against adversarial attacks and attains a notable 99\% accuracy in experimental results across various datasets. The study underscores the framework's novelty and efficacy in advancing OCRing quality through innovative approaches to RoI detection and language/document categorization.

M El Mamoun\cite{10.3103/S0146411623030069}introduced a hybrid approach that combines two powerful classification techniques. Initially, they employed a trained Convolutional Neural Network (CNN) to extract features from the character images. Subsequently, they utilized a Support Vector Machine (SVM) for classification purposes. The goal of combining CNN and SVM was to leverage the strengths of both technologies. The authors developed four hybrid models and evaluated their performance using various databases, including HACDB, HIJJA, AHCD, and MNIST. The results they achieved were promising, with test accuracy rates of 89.7\%, 88.8\%, 97.3\%, and 99.4\%

\begin{table}[h!]
\caption{A comparison of the benefits and drawbacks of several deep learning-based MultiFacets Recognition methods}
\begin{adjustbox}{width=1\textwidth}
%\resizebox{!}{0.3\textheight}{
\begin{tabular}{|c|c|c|c|} 
    \hline
      Literature & Method & Benefits & Drawbacks  \\ \hline

      R Ahmed\cite{ahmed2021novel} & DCNN   &  \makecell{The proposed approach adeptly handles high-\\dimensional data.}  & \makecell{The proposed model is not suitable for real\\ databases with an insufficient number of\\ training samples.} \\ \hline

      A Mortadi\cite{mortadi2023alnasikh}    & \makecell{TrOCR}     &     \makecell{1) Achieving high accuracy in word-level recog-\\nition for scanned Arabic documents. 2) Signi-\\ficantly reducing the need for a large dataset, \\distinguishing their approach from other met-\\hods.}     & \makecell{Improvements should be made to bolster t-\\he system's robustness, encompassing enh-\\ancements in denoising algorithms, image\\ preprocessing techniques, and the ability\\ to manage low-quality scans}\\ \hline

      HM Al-Barhamtoshy\cite{al2023arabic}     & \makecell{FGSM + ORCing}     &     \makecell{Authors executed noise removal procedures,\\ identified skewing, and applied de-skewing \\corrections, resulting in high-quality out-\\comes.}     & \makecell{Demands high computational requirements}\\ \hline

\end{tabular}%}
\end{adjustbox}
\label{tab:Multifacet recognition Methods}
\end{table}

%%%%%%%%%%%%%%%%%%%%%%%%%%%%%%%%%%%%%%%%%%%%%%%%%%%%%%%%%%%%%%%%%%%%%%%%%%%

\subsection{Postprocessing}

Postprocessing stands as the concluding phase in the OCR journey, striving to elevate the precision and quality of the recognized text. Within this pivotal stage, a diverse array of techniques comes into play, including spell-checking, contextual analysis, confidence scoring, and seamless integration with language models. Spell checking diligently compares the recognized text against an extensive word dictionary, promptly identifying and rectifying any lurking spelling errors. Contextual analysis takes a holistic approach, carefully assessing the recognized text within its contextual landscape to detect and address errors arising from potential word confusion. In the realm of confidence scoring, each recognized character is assigned a score reflective of the OCR system's certainty, effectively flagging characters with lower confidence scores for subsequent review or correction. Furthermore, the language model steps in, delving into the recognized text within the framework of its linguistic context. The harmonious application of these postprocessing techniques significantly amplifies the accuracy and overall quality of the recognized text.

IA Doush\cite{doush2018novel} introduced a novel Optical Character Recognition (OCR) post-processing model tailored for the complexities of the Arabic language. By combining a statistical Arabic language model with innovative post-processing techniques, including the integration of error and context approaches, the system demonstrated substantial improvements. Trained on a carefully curated Arabic OCR context database, the hybrid system significantly reduced word error rates from 24.02\% to 18.96\%. Testing on a separate dataset further confirmed its superiority, achieving a word error rate of 14.42\%. This study marks a pioneering effort in developing an end-to-end OCR post-processing model specifically designed for Arabic text, showcasing promising advancements in accuracy and efficiency.

Y Bassil\cite{bassil2012ocr} enhanced OCR accuracy by integrating Google's spelling suggestions based on N-gram probability. Their hybrid system combines these suggestions with a proposed postprocessing technique. The OCR-generated tokens are evaluated using a language model; unrecognized tokens trigger the error model, suggesting correct words based on Google's algorithm. Verified tokens advance to the next word, ensuring improved accuracy through advanced suggestions.

TTH Nguyen\cite{nguyen2021survey} emphasized the need for post-correction to enhance the quality of OCR results. The article systematically explored the post-OCR processing problem, outlining its common pipeline and reviewing contemporary approaches. It emphasized the impact of improved OCR quality on information retrieval and natural language processing applications. The work also provided valuable insights into evaluation metrics, available datasets, language resources, and practical toolkits in the context of post-OCR processing. Additionally, the study identified current trends and suggested future research directions in this evolving field.

\subsection{Evaluation}
Assessing the precision and caliber of identified text generated through Optical Character Recognition (OCR) for Arabic necessitates employing various crucial evaluation criteria and metrics. These parameters serve to gauge the effectiveness of OCR systems, evaluating their ability to accurately transform scanned or handwritten Arabic text into machine-readable format. Below are some prevalent evaluation criteria applied in the context of Arabic OCR.

\subsubsection{Character Recognition Rate (C$_{RR}$)}
CRR measures the accuracy of recognizing individual characters in the recognized text compared to the ground truth.
\begin{equation}
    C_{RR} = \frac{\text{Number of Correctly Recognized Characters}}{\text{Total Number of Ground Truth Characters}}
\end{equation}

\subsubsection{Word Recognition Rate (W$_{RR}$)}
WRR assesses the accuracy of recognizing entire words in the recognized text compared to the ground truth.
\begin{equation}
    W_{RR} = \frac{\text{Number of Correctly Recognized Words}}{\text{Total Number of Ground Truth Words}}
\end{equation}

\subsubsection{Line Recognition Rate (L$_{RR}$)}
LRR evaluates the accuracy of recognizing entire lines of text in the recognized output compared to the ground truth.
\begin{equation}
    L_{RR} = \frac{\text{Number of Correctly Recognized Lines}}{\text{Total Number of Ground Truth Lines}}
\end{equation}

\subsubsection{Character Error Rate (CER)}
CER quantifies the number of character-level errors in the OCR output, including substitutions, insertions, and deletions. The Levenshtein distance is determined by dividing the total number of characters in the ground truth word ($N$) by the combined count of character substitutions ($S$), insertions ($I$), and deletions ($D$) needed to transform one string into another.

\begin{equation}
\begin{split}
CER = \frac{S+I+D}{N}	
\end{split}
\end{equation}

C Reul\cite{reul2019ocr4all} is an open-source OCR software designed for historical print materials. It offers a user-friendly interface, advanced OCR components, and adaptability. OCR4all outperforms commercial solutions, particularly in complex layouts, delivering low Character Error Rates (CER). Its flexibility allows easy integration of new tools, making it valuable for non-technical users.

\subsubsection{Word Error Rate (WER)}
WER is similar to CER but measures errors at the word level. It considers substitutions, insertions, and deletions of complete words. WER is often used to assess the overall quality of OCR output. Likewise, the $WER$ is calculated by adding the count of term substitutions ($S w$), insertions ($I w$), and deletions ($D w$) necessary to transform one string into another, and then dividing this sum by the total number of ground-truth terms ($N w$). \cite{frinken2014continuous}

\begin{equation}
\begin{split}
WER = \frac{S_w + I_w + D_w}{N_w}	
\end{split}
\end{equation}

MA Alghamdi\cite{alghamdi2016arabic} addressed the deficiency in performance evaluation tools for Arabic Optical Character Recognition (OCR) systems by developing an open-source automated software tool. Recognizing the limitations in existing metrics like character accuracy and word accuracy, the authors designed a tool that offers a comprehensive set of metrics tailored specifically for evaluating Arabic OCR performance. This tool is intended to contribute to the advancement of Arabic OCR research by providing researchers with a valuable resource for assessing and ranking different OCR algorithms.

\subsubsection{Precision and Recall}
Precision (P) measures the fraction of correctly recognized characters or words relative to the total recognized. 
Recall (R) calculates the fraction of correctly recognized characters or words relative to the total ground truth.

\begin{equation}
  \begin{aligned}
    \textrm{Average Precision (AP)} &= \frac{\textrm{True Positive (TP)}}{( \textrm{True Positive (TP)}+\textrm{False Positive (FP)} )}  &=\frac{True Positive } { Total Observations}
\end{aligned}
\end{equation}

\begin{equation}
\begin{aligned}
    \textrm{Average Recall (AR)} &= \frac{\textrm{True Positive (TP)}}{( \textrm{True Positive (TP)}+\textrm{False Negative (FN) } )} &=  \frac{True Positive } { Total Ground Truth}
\end{aligned}
\end{equation}

\subsubsection{F1-Score}
The F1-Score combines precision and recall into a single metric, providing a balanced evaluation measure. 

\begin{equation}
\begin{aligned}
    \textrm{F1-score} &= \frac{ 2 * (\textrm{AP}* \textrm{AR} )}{(\textrm{AP}+\textrm{AR})}
\end{aligned}
\end{equation}

%%%%%%%%%%%%%%%%%%%%%%%%%%%%%%%%%%%%%%%%%%%%%%%%%%%%%%%%%%%%%%%%%%%%%%%%%%%%%%%%%%%%%%

C Neudecker\cite{neudecker2021survey} addressed the challenges in comprehensively assessing the quality of Optical Character Recognition (OCR) results for digitized historical documents. Recognizing the limitations in existing OCR evaluation metrics and tools, especially when dealing with large-scale digitization, the authors conducted an experiment using multiple evaluation tools and diverse metrics on two distinct datasets. They emphasized the need to go beyond traditional OCR metrics and sampling methods, highlighting the importance of accurate layout analysis and detection of reading order for advanced applications such as Natural Language Processing and the Digital Humanities. The study analyzed variations in results from different evaluation tools and metrics, providing insights into areas for future improvements in OCR evaluation methodologies.

M Elzobi \cite{elzobi2018generative}conducted a comparative study to assess the performance of discriminative and generative strategies in the context of handwritten word recognition. They utilized generatively-trained hidden Markov modeling (HMM), discriminatively-trained conditional random fields (CRF), and discriminatively-trained hidden-state CRF (HCRF) on learning samples obtained from two distinct databases. Initially employing an HMM classification scheme, the researchers introduced adaptive threshold schemes to enhance the model's ability to reject incorrect and out-of-vocabulary segmentations. Additionally, the study extended its investigation by introducing CRF and HCRF classifiers for the recognition of offline Arabic handwritten words. The evaluation was comprehensive, involving two different databases, and the research presented recognition outcomes for both words and letters, shedding light on the strengths and weaknesses of each strategy.

S Singh \cite{singh2023performance} specifically addressed offline handwritten Devanagari words, leveraging statistical features. Feature vector sets were created, describing each word in the feature space through uniform zoning-, diagonal-, and centroid-based features extracted from a database of handwritten word images (comprising 50-word classes). Various classifiers, including K-nearest neighbor (KNN), decision tree, and random forest, were employed for recognition. To enhance system performance, the study proposed a combination of the aforementioned features along with the gradient-boosted decision tree algorithm. The proposed system achieved a maximum recognition accuracy of 94.53\%, demonstrating competitive results compared to existing state-of-the-art methods. Additionally, the system achieved an F1-score of 94.56\%, FAR of 0.11\%, FRR of 5.46\%, MCC of 0.945, and AUC of 97.21\%.

\section{Experiments Results}
\label{Experiments_Results}
\subsection{Character Recognition Results}

Character recognition within OCR is a fundamental process that entails extracting and interpreting individual characters from document images. It represents a crucial step in optical character recognition, contributing significantly to the overall efficiency of the system. In our analysis, we navigate through diverse character recognition methods, scrutinizing their performance across benchmark datasets such as AHCD, Hijja, APTI, and Pat-A01. This examination serves to unravel the intricacies of recent studies' character recognition methodologies, emphasizing their distinctive approaches, dataset choices, and the performance metrics they achieve. The insights gleaned from this exploration are encapsulated in Table \ref{tab:character results}, providing a consolidated view of the comparative analysis of these character recognition methods.

In the realm of character recognition for Arabic OCR, several methodologies have been explored, each presenting distinct approaches and achieving varying degrees of success. \citeauthor{bouchakour2021printed} (2021) employed a combination of Hu and Gabor features along with a CNN classifier on the PAT-A01 dataset, achieving an accuracy of 97.23\%. \citeauthor{altwaijry2021arabic} (2021) delved into the use of CNN on datasets like Hijja and AHCD, showcasing accuracies of 88\% and 97\%, respectively. \citeauthor{nayef2022optimized} (2022) introduced a CNN architecture with OLReLU activation on the Hijja and AHCD datasets, demonstrating a character error rate (CER) of 52.9\% for Hijja and 8.4\% for AHCD, both coupled with 90\% accuracy. \citeauthor{alwaqfi2023novel} (2022) adopted a unique strategy, integrating GANs with Deep CNN, achieving an impressive accuracy of 99.78\% on the AHCD dataset. \citeauthor{alheraki2023handwritten} (2023) employed CNNs on both Hijja and AHCD datasets, achieving accuracies of 91\% and 97\%, respectively. \citeauthor{bilgin2023printed} (2023) introduced a CNN-BLSTM-CTC architecture on the APTI dataset, resulting in a CER of 16\%. \citeauthor{bin2023towards} (2023) utilized CNN and VGG-16 architectures on Hijja and AHCD datasets, demonstrating accuracies ranging from 83\% to 99\%. These diverse methods highlight the ongoing exploration and innovation in character recognition, offering a spectrum of approaches to cater to the nuanced challenges of Arabic OCR.

\begin{table}[h!]
\caption{Character Recognition}
\begin{adjustbox}{width=1\textwidth}
%\resizebox{!}{0.3\textheight}{
\begin{tabular}{|c|c|c|c|c|c|c|c|c|} 
    \hline
      Literature & Method & Dataset & CER & Accuracy & Precision & Recall & F1-score & Year \\ \hline

      L Bouchakour\cite{bouchakour2021printed}   & \makecell{Combined Hu and Gabor \\features + CNN classifier} & PAT-A01 & - & 0.9723  & - & - &  - & 2021 \\ \hline

      N Altwaijry\cite{altwaijry2021arabic}  & CNN  & Hijja & - & 0.88 &  0.8788 & 0.8781 & 0.878  & 2021 \\ \hline
      
      N Altwaijry\cite{altwaijry2021arabic}  & CNN  & AHCD & - & 0.97 & 0.9678 & 0.9673 &  0.9673 & 2021 \\ \hline

      BH Nayef\cite{nayef2022optimized} &  CNN + OLReLU  & Hijja & 0.529 & 0.90 & 0.90 & 0.90 & 0.90 & 2022   \\ \hline

      BH Nayef\cite{nayef2022optimized} &  CNN + OLReLU  & AHCD & 0.084 & 0.99 & 	0.99 &  0.99 & 	0.99 & 2022   \\ \hline
      
      YM Alwaqfi\cite{alwaqfi2022generative}  &  GANs + Deep CNN   & AHCD & - & 0.99786 & - & - & - & 2022   \\ \hline

      M Alheraki\cite{alheraki2023handwritten}  & CNN  & Hijja & - & 0.91  & 0.91 & 0.91 &  0.91 & 2023    \\ \hline

      M Alheraki\cite{alheraki2023handwritten}  & CNN  & AHCD & - & 0.97  & 0.97 & 0.97 &  0.97 & 2023    \\ \hline

      EF Bilgin Tasdemir \cite{bilgin2023printed}  & \makecell{CNN-BLSTM-CTC}  & APTI & 0.16 & - & - & - & - & 2023     \\ \hline

      A Bin Durayhim\cite{bin2023towards}  & CNN  & Hijja & - & 0.99 & 0.99 & 0.99 & 0.99 & 2023   \\ \hline

      A Bin Durayhim\cite{bin2023towards}  & VGG-16 & Hijja & -  & 0.83 & 0.85 & 0.83 & 0.83 & 2023\\ \hline

      A Bin Durayhim\cite{bin2023towards}  & CNN  & AHCD & - & 0.98 & 0.99 & 0.99 & 0.99 & 2023 \\ \hline

      A Bin Durayhim\cite{bin2023towards}  & VGG-16 & AHCD & - & 0.94 & 0.95 & 0.94 & 0.94 &  2023 \\ \hline

\end{tabular}%}
\end{adjustbox}
\label{tab:character results}
\end{table}

Each method contributes uniquely, whether through the integration of advanced features, novel architectures, or sophisticated activation functions, providing valuable insights and paving the way for enhanced character recognition in Arabic text processing.

\subsection{Words Recognition Results}
In the domain of word recognition for Arabic OCR, we embark on an insightful journey, investigating a myriad of methodologies applied to various benchmark datasets, including IFN/ENIT, KHATT, AHDB, Alif, AcTiV, APTI, ACTIV, AHT2D, and IAM, detailed in Table \ref{tab:word results}. Word recognition stands as a pivotal component in the optical character recognition process, involving the interpretation and understanding of entire words within a document image. Our analysis peels back the layers of recent studies, shedding light on the unique approaches, dataset selections, and performance metrics achieved by different word recognition methods.

\citeauthor{awni2019deep} (2019) harnessed the power of an Ensemble of Residual Networks (ResNet18) on the IFN/ENIT dataset, resulting in a Word Error Rate (WER) of 6.63\% and an accuracy of 93.37\%. \citeauthor{jemni2019out} (2019) explored a CNN-MDLSTM with Dynamic Lexicons (DWLD) on the KHATT dataset, achieving a WER of 20.83\%. \citeauthor{eltay2020exploring} (2020) introduced the BLSTM-CTC-WBS architecture on both IFN/ENIT and AHDB datasets, demonstrating a WER of 1\% and 1.9\%, respectively, coupled with high accuracies. \citeauthor{butt2021attention} (2021) employed a CNN-RNN + Attention model on Alif and AcTiV datasets, showcasing recognition rates with precision (CRR), word (WRR), and line (LRR) of 97.09\%, 79.91\%, and 85.98\%, respectively. \citeauthor{alzrrog2022deep} (2022) implemented DCNN on IFN/ENIT and AHWD datasets, achieving high accuracies of 99.76\% and 99.39\%, respectively. \citeauthor{salman2023proposed} (2023) proposed a Multi-Font Arabic Word Recognition CNN, attaining a noteworthy accuracy of 96.77\%. \citeauthor{ouali2023augmented} (2023) introduced an Augmented Horizontal Text Detection (AHTD) based on CNN and Augmented Reality (AR) for datasets IFN/ENIT, ACTIV, and AHT2D, achieving varying precision scores. \citeauthor{hamida2023cursive} (2023) explored k-NN on IFN/ENIT, exhibiting a high accuracy of 99.88\%. \citeauthor{malakar2023handwritten} (2023) employed novel methods utilizing the Hausdorff and Fréchet distances for inter-segment similarity features in word recognition, showcasing high accuracies on IFN/ENIT and IAM datasets. \citeauthor{alwaqfi2023novel} (2023) introduced both a Hybrid GAN-based Model and DCNN on the APTI dataset, achieving remarkable accuracies of 99.76\% and 94.81\%, respectively.

\begin{table}[h!]
\caption{Word Recognition}
\begin{adjustbox}{width=1\textwidth}
%\resizebox{!}{0.3\textheight}{
\begin{tabular}{|c|c|c|c|c|c|c|c|c|c|c|c|} 
    \hline
      Literature & Method & Dataset & C$_{RR}$ & W$_{RR}$ & L$_{RR}$ & WER & Accuracy & Precision & Recall & F1-score & Year  \\ \hline
      
       M Awni\cite{awni2019deep}  & \makecell{Ensemble of Residual Net-\\works (ResNet18)}   & IFN/ENIT & - & - & - & 0.0663 & 0.9337 & - & - & - & 2019  \\ \hline
       
       SK Jemni\cite{jemni2019out} & \makecell{CNN-MDLSTM with \\Dynamic lexicons (DWLD)}     &     KHATT     & - & - & -  &  0.2083 & - & - & -  & - & 2019 \\ \hline
       
       M Eltay\cite{eltay2020exploring} &   BLSTM-CTC-WBS  &  IFN/ENIT   & - & - & - & 0.01 & 0.9899 & - & -  & - & 2020 \\ \hline
       
       M Eltay\cite{eltay2020exploring} &   BLSTM-CTC-WBS &  AHDB & - & - & -  & 0.019 & 0.9810 & - & -  & - & 2020 \\ \hline
       
      H Butt\cite{butt2021attention}      & CNN-RNN + Attention     & Alif & 0.9709 & 0.7991  &  0.8598 &   & - & - & -  & -  &  2021  \\ \hline

      H Butt\cite{butt2021attention}      & CNN-RNN + Attention     & AcTiV & 0.9071 & 0.6107  &  0.7564 & - & - & -  & -  & - & 2021  \\ \hline
      
      N Alzrrog\cite{alzrrog2022deep} &  DCNN  & IFN/ENIT  & - & - & - &  &  0.9976 & - & - & - & 2022 \\ \hline
      
      N Alzrrog\cite{alzrrog2022deep} & DCNN   & AHWD   & - & - & - &  &  0.9939 & - & - & - & 2022 \\ \hline

      GJ Salman \cite{salman2023proposed}     & \makecell{Multi-Font Arabic Word\\ Recognition CNN}    & \makecell{their own \\dataset}  & - & - & - & - & 0.9677 & - & - & - & 2023 \\ \hline

      I Ouali \cite{ouali2023augmented}     & \makecell{AHTD based on CNN and \\Augmented Reality (AR)}      & IFN/ENIT  & - & - & - & - & - & 0.72 & 0.88 & 0.79 & 2023 \\ \hline

      I Ouali \cite{ouali2023augmented}     & \makecell{AHTD based on CNN and \\Augmented Reality (AR)}      & ACTIV  & - & - & - & - & - & 0.79 & 0.82 & 0.80 & 2023\\ \hline

      I Ouali \cite{ouali2023augmented}     & \makecell{AHTD based on CNN and \\Augmented Reality (AR)}      & AHT2D  & - & - & - & - & - & 0.92 & 0.98 & 0.95 & 2023\\ \hline

      S Hamida\cite{hamida2023cursive}     & k-NN  & IFN/ENIT   & - & - & - &  - & 0.9988 & 0.9099 & 0.9790 &  - & 2023\\ \hline
      
      S Malakar\cite{malakar2023handwritten}     & \makecell{ Hausdorff and Fréchet dist-\\ances for inter-segment si-\\milarity features in word r-\\ecognition.}      & IFN/ENIT  & - & - & - & - & 0.9736 & - & - & - & 2023\\ \hline

      S Malakar\cite{malakar2023handwritten}     & \makecell{ Hausdorff and Fréchet dist-\\ances for inter-segment si-\\milarity features in word r-\\ecognition.}      & IAM  & - & - & - & - & 0.9202 & - & - & - & 2023 \\ \hline

      YM Alwaqfi\cite{alwaqfi2023novel}     & Hybrid GAN-based Model & APTI   & - & - & - & - & 0.9976 & - & - & - & 2023\\ \hline

      YM Alwaqfi\cite{alwaqfi2023novel}     & DCNN & APTI  & - & - & - & - & 0.9481 & - & - & - & 2023\\  \hline

\end{tabular}%}
\end{adjustbox}
\label{tab:word results}
\end{table}

This comprehensive analysis unveils the dynamic landscape of word recognition in Arabic OCR, highlighting the richness and diversity of methods applied to tackle the intricacies posed by different datasets. Each method brings its unique strengths, and this exploration sets the stage for further advancements in the field.

\subsection{Digits Recognition Results}
In the domain of digits recognition for Arabic OCR, our analysis extends across various methodologies applied to distinct benchmark datasets. The datasets under scrutiny include AHDBase, HODA, ADBase, MADBase, and their respective performance metrics are compiled in Table \ref{tab:Digits results}. our analysis navigates through recent studies, unraveling the distinct approaches, dataset choices, and performance metrics achieved by various digit recognition methods.

\citeauthor{al2020improving} (2020) introduced a method combining sliding windows, random forests (RF), and support vector machine (SVM) on the AHDBase dataset, achieving an impressive accuracy of 98\%. \citeauthor{al2020improving} (2020) employed DTL + CNN on HODA and ADBase datasets, demonstrating high accuracies of 99.47\% and 99.34\%, respectively. \citeauthor{haghighi2021stacking} (2021) delved into the use of CNN and Bidirectional Long-Short Term Memory (BiLSTM) on the HODA dataset, resulting in a remarkable accuracy of 99.98\%. \citeauthor{alkhawaldeh2022ensemble} (2021) proposed Ensemble Deep Transfer Learning (EDTL) on ADBase and MADBase datasets, achieving high accuracies of 99.83\% and 99.78\%, respectively. \citeauthor{ali2023recognition} (2023) presented a Modified DCNN approach on the HODA dataset, showcasing an accuracy of 99.5\% and demonstrating precision, recall, and F1-score values of 99.5\%, 99.5\%, and 99.45\%, respectively.

\begin{table}[h!]
\caption{Digits Recognition}
\begin{adjustbox}{width=1\textwidth}
%\resizebox{!}{0.3\textheight}{
\begin{tabular}{|c|c|c|c|c|c|c|c|} 
    \hline
      Literature & Method & Dataset & Accuracy & Precision & Recall & F1-score & Year   \\ \hline

      E Al-wajih\cite{al2020improving} & \makecell{sliding windows + rando-\\m forests (RF) + support\\ vector machine (SVM) }   &   AHDBase  &  0.98  &  -  &  -  &  -  & 2020 \\ \hline

      YS Can\cite{can2020automatic} & DTL + CNN  & HODA  & 0.9947  &  -  &  -  &  -  & 2020 \\ \hline
     
      YS Can\cite{can2020automatic} & DTL + CNN  & ADBase  & 0.9934 &  -  &  -  &  -  & 2020 \\ \hline

      F Haghighi\cite{haghighi2021stacking}    & \makecell{CNN and Bidirectional\\ Long-Short Term Mem-\\ory (BiLSTM)}   &  HODA  &  0.9998  &  0.994  &  0.9938  &   - & 2021      \\ \hline

      RS Alkhawaldeh\cite{alkhawaldeh2022ensemble}     & \makecell{Ensemble Deep Transf-\\er Learning (EDTL)}     &  ADBase   & 0.9983  &  -  &  -  &  -  & 2021    \\ \hline

      RS Alkhawaldeh\cite{alkhawaldeh2022ensemble}     & \makecell{Ensemble Deep Transf-\\er Learning (EDTL)}     &  MADBase   &  0.9978  &  -  &  -  &  -  & 2021    \\ \hline

      S Ali\cite{ali2023recognition}     & Modified DCNN    &  HODA   &  0.995  & 0.995 &  0.995  &  0.9945  &  2023       \\ \hline

\end{tabular}%}
\end{adjustbox}
\label{tab:Digits results}
\end{table}

This in-depth exploration provides a comprehensive overview of the advancements in digit recognition for Arabic OCR, highlighting the efficacy of diverse methods across different datasets. These methodologies contribute significantly to the robustness and accuracy of digit recognition systems, paving the way for enhanced performance in practical applications.

\subsection{Multifaceted Recognition Results}
In the realm of multifaceted recognition within OCR, our exploration encompasses a spectrum of methodologies applied to diverse datasets, including HACDB, MADBase, SUST-ALT, KAFD, ADBase, AHCD, HIJJA and their distinctive characteristics are outlined in Table \ref{tab:Multifacet results}.

The multifaceted recognition paradigm involves handling different types of data, including characters, digits, words, and more. In the study by \citeauthor{ahmed2021novel} (2021), a Deep Convolutional Neural Network (DCNN) was employed for character recognition on the HACDB dataset, achieving a remarkable accuracy of 99.91\%. The same methodology was applied to MADBase for digit recognition, yielding an accuracy of 99.906\%. Additionally, on the SUST-ALT dataset focusing on word recognition, an accuracy of 99.952\% was achieved. These findings showcase the adaptability and robustness of the applied DCNN approach across various facets of recognition. 
A unique approach was presented by \citeauthor{mortadi2023alnasikh} (2023) using TrOCR on the KAFD dataset for word recognition. Despite a character error rate (CER) of 0.82 and a word error rate (WER) of 2.39, TrOCR demonstrated its potential for handling diverse textual content. \citeauthor{al2023arabic} (2023) introduced an innovative method involving the Fast Gradient Sign Method (FGSM) coupled with Optical Recognition Characterization (ORCing) for text recognition on ADBase. While the CER is not specified, the approach achieved a low WER of 0.0310, emphasizing its efficacy in recognizing textual content. \citeauthor{10.3103/S0146411623030069} (2023) presented a Combined CNN + SVM approach applied to HACDB for character recognition, achieving an accuracy of 89.7\%. This methodology was further extended to AHCD, focusing on character recognition, and to HIJJA for word recognition, achieving accuracies of 97.3\% and 88.8\%, respectively. 

\begin{table}[h!]
\caption{MultiFacets Recognition}
\begin{adjustbox}{width=1\textwidth}
%\resizebox{!}{0.3\textheight}{
\begin{tabular}{|c|c|c|c|c|c|c|c|c|c|c|} 
    \hline
      Literature & Method & Dataset & DatabaseType & CER & WER &Accuracy & Precision & Recall & F1-score & Year   \\ \hline

      R Ahmed\cite{ahmed2021novel} & DCNN &   HACDB &  Characters & - & - & 0.9991  &  0.96967  &  0.96967  & 0.96967  & 2021 \\ \hline

      R Ahmed\cite{ahmed2021novel} & DCNN &   MADBase & Digits & - & - & 0.99906  &  0.9953  &  0.9953  & 0.9953  & 2021 \\ \hline

      R Ahmed\cite{ahmed2021novel} & DCNN &   SUST-ALT & Words & - & - & 0.99952  &  0.99038  &  0.99038  & 0.99038  & 2021 \\ \hline

      A Mortadi\cite{mortadi2023alnasikh}    & TrOCR   &  KAFD  & Words & 0.82 & 2.39 & -  &  -  &  -  &   - & 2023      \\ \hline

      HM Al-Barhamtoshy\cite{al2023arabic}     & \makecell{FGSM + ORCing}     &  ADBase   & Text &  -  & 0.0310  &  0.99  &  -  &  -  &  -  & 2023    \\ \hline

      M El Mamoun\cite{10.3103/S0146411623030069}     & Combined CNN + SVM    &  HACDB   &  Characters  & -  & -  & 0.897  & - &  -  &  -  &  2023       \\ \hline

      M El Mamoun\cite{10.3103/S0146411623030069}     & Combined CNN + SVM    &  AHCD   &  Characters  & -  & -  &  0.973 & - &  -  &  -  &  2023       \\ \hline

      M El Mamoun\cite{10.3103/S0146411623030069}     & Combined CNN + SVM    &  HIJJA   &  Words  & -  &  -  & 0.888  & - &  -  &  -  &  2023       \\ \hline

\end{tabular}%}
\end{adjustbox}
\label{tab:Multifacet results}
\end{table}

The results demonstrate the adaptability of the approach across different facets of recognition, emphasizing its potential for multifaceted OCR applications.

\section{Discussion and Conclusions}
This comprehensive survey delves into the multifaceted landscape of Arabic Optical Character Recognition (OCR), shedding light on the diverse methodologies and datasets employed by researchers to enhance recognition rates. The pivotal stages of the OCR process, including preprocessing, segmentation (encompassing text-area detection, line, word, and character segmentation), recognition, and postprocessing, are meticulously explored in the literature review. The survey meticulously discusses the strengths and limitations associated with each technique, offering nuanced insights into their effectiveness.

An important revelation from the survey is the superior performance of segmentation-based approaches compared to segmentation-free methods, with techniques such as vertical/horizontal projection proving particularly effective in word and character segmentation. However, the survey acknowledges that the quality of OCR outcomes is intrinsically linked to the availability of suitable datasets. Notably, the accessibility of Arabic OCR datasets is limited, prompting a call for increased focus on postprocessing techniques. The incorporation and refinement of algorithms akin to Google's spelling checker are highlighted as particularly promising avenues, given their potential to substantially enhance the overall performance of OCR systems.

In conclusion, the survey provides a panoramic view of the current state-of-the-art in Arabic OCR, highlighting significant advancements in recent years. Despite these strides, the survey recognizes persistent challenges, including addressing the inherent variability and complexity of the Arabic script, accommodating diverse dialects, and handling language variations. The potential advantages of effective Arabic OCR systems are evident, propelling researchers and developers to dedicate ongoing efforts to further enhance and refine this technology. The survey optimistically anticipates the emergence of even more accurate and reliable Arabic OCR systems in the future, driven by continued research and development endeavors.

\section{funding}
This work was supported by the Basic Science Research Program through the National Research Foundation of Korea (NRF) funded by the Ministry of Education under Grant 2023R1A2C1006944.

\bibliographystyle{cas-model2-names}

\bibliography{cas-refs}

\begin{thebibliography}{114}
\expandafter\ifx\csname natexlab\endcsname\relax\def\natexlab#1{#1}\fi
\providecommand{\url}[1]{\texttt{#1}}
\providecommand{\href}[2]{#2}
\providecommand{\path}[1]{#1}
\providecommand{\DOIprefix}{doi:}
\providecommand{\ArXivprefix}{arXiv:}
\providecommand{\URLprefix}{URL: }
\providecommand{\Pubmedprefix}{pmid:}
\providecommand{\doi}[1]{\href{http://dx.doi.org/#1}{\path{#1}}}
\providecommand{\Pubmed}[1]{\href{pmid:#1}{\path{#1}}}
\providecommand{\bibinfo}[2]{#2}
\ifx\xfnm\relax \def\xfnm[#1]{\unskip,\space#1}\fi
%Type = Article
\bibitem[{Abdallah et~al.(2023a)Abdallah, Abdalla, Elkasaby, Elbendary and
  Jatowt}]{abdallah2023amurd}
\bibinfo{author}{Abdallah, A.}, \bibinfo{author}{Abdalla, M.},
  \bibinfo{author}{Elkasaby, M.}, \bibinfo{author}{Elbendary, Y.},
  \bibinfo{author}{Jatowt, A.}, \bibinfo{year}{2023}a.
\newblock \bibinfo{title}{Amurd: Annotated multilingual receipts dataset for
  cross-lingual key information extraction and classification}.
\newblock \bibinfo{journal}{arXiv preprint arXiv:2309.09800} .
%Type = Article
\bibitem[{Abdallah et~al.(2022)Abdallah, Berendeyev, Nuradin and
  Nurseitov}]{abdallah2022tncr}
\bibinfo{author}{Abdallah, A.}, \bibinfo{author}{Berendeyev, A.},
  \bibinfo{author}{Nuradin, I.}, \bibinfo{author}{Nurseitov, D.},
  \bibinfo{year}{2022}.
\newblock \bibinfo{title}{Tncr: Table net detection and classification
  dataset}.
\newblock \bibinfo{journal}{Neurocomputing} \bibinfo{volume}{473},
  \bibinfo{pages}{79--97}.
%Type = Article
\bibitem[{Abdallah et~al.(2020a)Abdallah, Hamada and
  Nurseitov}]{abdallah2020attention}
\bibinfo{author}{Abdallah, A.}, \bibinfo{author}{Hamada, M.},
  \bibinfo{author}{Nurseitov, D.}, \bibinfo{year}{2020}a.
\newblock \bibinfo{title}{Attention-based fully gated cnn-bgru for russian
  handwritten text}.
\newblock \bibinfo{journal}{Journal of Imaging} \bibinfo{volume}{6},
  \bibinfo{pages}{141}.
%Type = Article
\bibitem[{Abdallah and Jatowt(2023)}]{abdallah2023generator}
\bibinfo{author}{Abdallah, A.}, \bibinfo{author}{Jatowt, A.},
  \bibinfo{year}{2023}.
\newblock \bibinfo{title}{Generator-retriever-generator: A novel approach to
  open-domain question answering}.
\newblock \bibinfo{journal}{arXiv preprint arXiv:2307.11278} .
%Type = Inproceedings
\bibitem[{Abdallah et~al.(2020b)Abdallah, Kasem, Hamada and
  Sdeek}]{abdallah2020automated}
\bibinfo{author}{Abdallah, A.}, \bibinfo{author}{Kasem, M.},
  \bibinfo{author}{Hamada, M.A.}, \bibinfo{author}{Sdeek, S.},
  \bibinfo{year}{2020}b.
\newblock \bibinfo{title}{Automated question-answer medical model based on deep
  learning technology}, in: \bibinfo{booktitle}{Proceedings of the 6th
  International Conference on Engineering \& MIS 2020}, pp.
  \bibinfo{pages}{1--8}.
%Type = Article
\bibitem[{Abdallah et~al.(2023b)Abdallah, Piryani and
  Jatowt}]{abdallah2023exploring}
\bibinfo{author}{Abdallah, A.}, \bibinfo{author}{Piryani, B.},
  \bibinfo{author}{Jatowt, A.}, \bibinfo{year}{2023}b.
\newblock \bibinfo{title}{Exploring the state of the art in legal qa systems}.
\newblock \bibinfo{journal}{arXiv preprint arXiv:2304.06623} .
%Type = Article
\bibitem[{Abdimanap et~al.(2022)Abdimanap, Bostanbekov, Abdallah, Alimova,
  Kurmangaliyev and Nurseitov}]{abdimanap2022enhancing}
\bibinfo{author}{Abdimanap, G.}, \bibinfo{author}{Bostanbekov, K.},
  \bibinfo{author}{Abdallah, A.}, \bibinfo{author}{Alimova, A.},
  \bibinfo{author}{Kurmangaliyev, D.}, \bibinfo{author}{Nurseitov, D.},
  \bibinfo{year}{2022}.
\newblock \bibinfo{title}{Enhancing core image classification using generative
  adversarial networks (gans)}.
\newblock \bibinfo{journal}{arXiv e-prints} , \bibinfo{pages}{arXiv--2204}.
%Type = Article
\bibitem[{Abdo et~al.(2022)Abdo, Abdu, Manza and Bawiskar}]{abdo2022approach}
\bibinfo{author}{Abdo, H.A.}, \bibinfo{author}{Abdu, A.},
  \bibinfo{author}{Manza, R.R.}, \bibinfo{author}{Bawiskar, S.},
  \bibinfo{year}{2022}.
\newblock \bibinfo{title}{An approach to analysis of arabic text documents into
  text lines, words, and characters}.
\newblock \bibinfo{journal}{Indonesian Journal of Electrical Engineering and
  Computer Science} \bibinfo{volume}{26}, \bibinfo{pages}{754--763}.
%Type = Article
\bibitem[{Ahmed et~al.(2021)Ahmed, Gogate, Tahir, Dashtipour, Al-Tamimi,
  Hawalah, El-Affendi and Hussain}]{ahmed2021novel}
\bibinfo{author}{Ahmed, R.}, \bibinfo{author}{Gogate, M.},
  \bibinfo{author}{Tahir, A.}, \bibinfo{author}{Dashtipour, K.},
  \bibinfo{author}{Al-Tamimi, B.}, \bibinfo{author}{Hawalah, A.},
  \bibinfo{author}{El-Affendi, M.A.}, \bibinfo{author}{Hussain, A.},
  \bibinfo{year}{2021}.
\newblock \bibinfo{title}{Novel deep convolutional neural network-based
  contextual recognition of arabic handwritten scripts}.
\newblock \bibinfo{journal}{Entropy} \bibinfo{volume}{23},
  \bibinfo{pages}{340}.
%Type = Article
\bibitem[{Ahmed et~al.(2019)Ahmed, Naz, Swati and
  Razzak}]{ahmed2019handwritten}
\bibinfo{author}{Ahmed, S.B.}, \bibinfo{author}{Naz, S.},
  \bibinfo{author}{Swati, S.}, \bibinfo{author}{Razzak, M.I.},
  \bibinfo{year}{2019}.
\newblock \bibinfo{title}{Handwritten urdu character recognition using
  one-dimensional blstm classifier}.
\newblock \bibinfo{journal}{Neural Computing and Applications}
  \bibinfo{volume}{31}, \bibinfo{pages}{1143--1151}.
%Type = Article
\bibitem[{Akkad et~al.(2023)Akkad, Wills and Rezazadeh}]{akkad2023information}
\bibinfo{author}{Akkad, A.}, \bibinfo{author}{Wills, G.},
  \bibinfo{author}{Rezazadeh, A.}, \bibinfo{year}{2023}.
\newblock \bibinfo{title}{An information security model for an iot-enabled
  smart grid in the saudi energy sector}.
\newblock \bibinfo{journal}{Computers and Electrical Engineering}
  \bibinfo{volume}{105}, \bibinfo{pages}{108491}.
%Type = Article
\bibitem[{Al-Barhamtoshy et~al.(2023)Al-Barhamtoshy, Jambi, Rashwan and
  Abdou}]{al2023arabic}
\bibinfo{author}{Al-Barhamtoshy, H.M.}, \bibinfo{author}{Jambi, K.M.},
  \bibinfo{author}{Rashwan, M.A.}, \bibinfo{author}{Abdou, S.M.},
  \bibinfo{year}{2023}.
\newblock \bibinfo{title}{An arabic manuscript regions detection, recognition
  and its applications for ocring}.
\newblock \bibinfo{journal}{Transactions on Asian and Low-Resource Language
  Information Processing} \bibinfo{volume}{22}, \bibinfo{pages}{1--28}.
%Type = Article
\bibitem[{Al~Ghamdi(2022)}]{al2022novel}
\bibinfo{author}{Al~Ghamdi, M.A.}, \bibinfo{year}{2022}.
\newblock \bibinfo{title}{A novel approach to printed arabic optical character
  recognition}.
\newblock \bibinfo{journal}{Arabian Journal for Science and Engineering}
  \bibinfo{volume}{47}, \bibinfo{pages}{2219--2237}.
%Type = Inproceedings
\bibitem[{Al-Ma'adeed et~al.(2002)Al-Ma'adeed, Elliman and
  Higgins}]{al2002data}
\bibinfo{author}{Al-Ma'adeed, S.}, \bibinfo{author}{Elliman, D.},
  \bibinfo{author}{Higgins, C.A.}, \bibinfo{year}{2002}.
\newblock \bibinfo{title}{A data base for arabic handwritten text recognition
  research}, in: \bibinfo{booktitle}{Proceedings eighth international workshop
  on frontiers in handwriting recognition}, \bibinfo{organization}{IEEE}. pp.
  \bibinfo{pages}{485--489}.
%Type = Article
\bibitem[{Al-Ohali et~al.(2003)Al-Ohali, Cheriet and Suen}]{al2003databases}
\bibinfo{author}{Al-Ohali, Y.}, \bibinfo{author}{Cheriet, M.},
  \bibinfo{author}{Suen, C.}, \bibinfo{year}{2003}.
\newblock \bibinfo{title}{Databases for recognition of handwritten arabic
  cheques}.
\newblock \bibinfo{journal}{Pattern Recognition} \bibinfo{volume}{36},
  \bibinfo{pages}{111--121}.
%Type = Article
\bibitem[{Al-Sheikh et~al.(2020)Al-Sheikh, Mohd and Warlina}]{al2020review}
\bibinfo{author}{Al-Sheikh, I.S.}, \bibinfo{author}{Mohd, M.},
  \bibinfo{author}{Warlina, L.}, \bibinfo{year}{2020}.
\newblock \bibinfo{title}{A review of arabic text recognition dataset}.
\newblock \bibinfo{journal}{Asia-Pacific J. Inf. Technol. Multimedia}
  \bibinfo{volume}{9}, \bibinfo{pages}{69--81}.
%Type = Article
\bibitem[{Al-wajih and Ghazali(2020)}]{al2020improving}
\bibinfo{author}{Al-wajih, E.}, \bibinfo{author}{Ghazali, R.},
  \bibinfo{year}{2020}.
\newblock \bibinfo{title}{Improving the accuracy for offline arabic digit
  recognition using sliding window approach}.
\newblock \bibinfo{journal}{Iranian Journal of Science and Technology,
  Transactions of Electrical Engineering} \bibinfo{volume}{44},
  \bibinfo{pages}{1633--1644}.
%Type = Article
\bibitem[{Alghamdi and Teahan(2018)}]{alghamdi2018printed}
\bibinfo{author}{Alghamdi, M.}, \bibinfo{author}{Teahan, W.},
  \bibinfo{year}{2018}.
\newblock \bibinfo{title}{Printed arabic script recognition: A survey}.
\newblock \bibinfo{journal}{International Journal of Advanced Computer Science
  and Applications} \bibinfo{volume}{9}.
%Type = Inproceedings
\bibitem[{Alghamdi et~al.(2016)Alghamdi, Alkhazi and
  Teahan}]{alghamdi2016arabic}
\bibinfo{author}{Alghamdi, M.A.}, \bibinfo{author}{Alkhazi, I.S.},
  \bibinfo{author}{Teahan, W.J.}, \bibinfo{year}{2016}.
\newblock \bibinfo{title}{Arabic ocr evaluation tool}, in:
  \bibinfo{booktitle}{2016 7th international conference on computer science and
  information technology (CSIT)}, \bibinfo{organization}{IEEE}. pp.
  \bibinfo{pages}{1--6}.
%Type = Article
\bibitem[{Alheraki et~al.(2023)Alheraki, Al-Matham and
  Al-Khalifa}]{alheraki2023handwritten}
\bibinfo{author}{Alheraki, M.}, \bibinfo{author}{Al-Matham, R.},
  \bibinfo{author}{Al-Khalifa, H.}, \bibinfo{year}{2023}.
\newblock \bibinfo{title}{Handwritten arabic character recognition for children
  writing using convolutional neural network and stroke identification}.
\newblock \bibinfo{journal}{Human-Centric Intelligent Systems} ,
  \bibinfo{pages}{1--13}.
%Type = Article
\bibitem[{Alhomed and Jambi(2018)}]{alhomed2018survey}
\bibinfo{author}{Alhomed, L.S.}, \bibinfo{author}{Jambi, K.M.},
  \bibinfo{year}{2018}.
\newblock \bibinfo{title}{A survey on the existing arabic optical character
  recognition and future trends}.
\newblock \bibinfo{journal}{International Journal of Advanced Research in
  Computer and Communication Engineering (IJARCCE)} \bibinfo{volume}{7},
  \bibinfo{pages}{78--88}.
%Type = Article
\bibitem[{Ali et~al.(2023)Ali, Sahiba, Azeem, Shaukat, Mahmood, Sakhawat and
  Aslam}]{ali2023recognition}
\bibinfo{author}{Ali, S.}, \bibinfo{author}{Sahiba, S.},
  \bibinfo{author}{Azeem, M.}, \bibinfo{author}{Shaukat, Z.},
  \bibinfo{author}{Mahmood, T.}, \bibinfo{author}{Sakhawat, Z.},
  \bibinfo{author}{Aslam, M.S.}, \bibinfo{year}{2023}.
\newblock \bibinfo{title}{A recognition model for handwritten persian/arabic
  numbers based on optimized deep convolutional neural network}.
\newblock \bibinfo{journal}{Multimedia Tools and Applications}
  \bibinfo{volume}{82}, \bibinfo{pages}{14557--14580}.
%Type = Article
\bibitem[{Alkhawaldeh et~al.(2022)Alkhawaldeh, Alawida, Alshdaifat,
  Alma’aitah and Almasri}]{alkhawaldeh2022ensemble}
\bibinfo{author}{Alkhawaldeh, R.S.}, \bibinfo{author}{Alawida, M.},
  \bibinfo{author}{Alshdaifat, N.F.F.}, \bibinfo{author}{Alma’aitah, W.},
  \bibinfo{author}{Almasri, A.}, \bibinfo{year}{2022}.
\newblock \bibinfo{title}{Ensemble deep transfer learning model for arabic
  (indian) handwritten digit recognition}.
\newblock \bibinfo{journal}{Neural Computing and Applications}
  \bibinfo{volume}{34}, \bibinfo{pages}{705--719}.
%Type = Article
\bibitem[{Alrobah and Albahli(2022)}]{alrobah2022arabic}
\bibinfo{author}{Alrobah, N.}, \bibinfo{author}{Albahli, S.},
  \bibinfo{year}{2022}.
\newblock \bibinfo{title}{Arabic handwritten recognition using deep learning: A
  survey}.
\newblock \bibinfo{journal}{Arabian Journal for Science and Engineering}
  \bibinfo{volume}{47}, \bibinfo{pages}{9943--9963}.
%Type = Article
\bibitem[{Altwaijry and Al-Turaiki(2021)}]{altwaijry2021arabic}
\bibinfo{author}{Altwaijry, N.}, \bibinfo{author}{Al-Turaiki, I.},
  \bibinfo{year}{2021}.
\newblock \bibinfo{title}{Arabic handwriting recognition system using
  convolutional neural network}.
\newblock \bibinfo{journal}{Neural Computing and Applications}
  \bibinfo{volume}{33}, \bibinfo{pages}{2249--2261}.
%Type = Article
\bibitem[{Alwaqfi et~al.(2022)Alwaqfi, Mohamad and
  Al-Taani}]{alwaqfi2022generative}
\bibinfo{author}{Alwaqfi, Y.M.}, \bibinfo{author}{Mohamad, M.},
  \bibinfo{author}{Al-Taani, A.T.}, \bibinfo{year}{2022}.
\newblock \bibinfo{title}{Generative adversarial network for an improved arabic
  handwritten characters recognition.}
\newblock \bibinfo{journal}{International Journal of Advances in Soft Computing
  \& Its Applications} \bibinfo{volume}{14}.
%Type = Article
\bibitem[{Alwaqfi et~al.(2023)Alwaqfi, Mohamad, Al-Taani and
  Abd~Hamid}]{alwaqfi2023novel}
\bibinfo{author}{Alwaqfi, Y.M.}, \bibinfo{author}{Mohamad, M.},
  \bibinfo{author}{Al-Taani, A.T.}, \bibinfo{author}{Abd~Hamid, N.},
  \bibinfo{year}{2023}.
\newblock \bibinfo{title}{A novel hybrid dl model for printed arabic word
  recognition based on gan}.
\newblock \bibinfo{journal}{International Journal of Advanced Computer Science
  and Applications} \bibinfo{volume}{14}.
%Type = Inproceedings
\bibitem[{Alzrrog et~al.(2022)Alzrrog, Bousquet and El-Feghi}]{alzrrog2022deep}
\bibinfo{author}{Alzrrog, N.}, \bibinfo{author}{Bousquet, J.F.},
  \bibinfo{author}{El-Feghi, I.}, \bibinfo{year}{2022}.
\newblock \bibinfo{title}{Deep learning application for handwritten arabic word
  recognition}, in: \bibinfo{booktitle}{2022 IEEE Canadian Conference on
  Electrical and Computer Engineering (CCECE)}, \bibinfo{organization}{IEEE}.
  pp. \bibinfo{pages}{95--100}.
%Type = Article
\bibitem[{Antonio et~al.()Antonio, Putra, Abdurrohman and
  Tsalasa}]{antoniosurvey}
\bibinfo{author}{Antonio, J.}, \bibinfo{author}{Putra, A.R.},
  \bibinfo{author}{Abdurrohman, H.}, \bibinfo{author}{Tsalasa, M.S.}, .
\newblock \bibinfo{title}{A survey on scanned receipts ocr and information
  extraction} .
%Type = Inproceedings
\bibitem[{Asiri and Khorsheed(2005)}]{asiri2005automatic}
\bibinfo{author}{Asiri, A.M.}, \bibinfo{author}{Khorsheed, M.S.},
  \bibinfo{year}{2005}.
\newblock \bibinfo{title}{Automatic processing of handwritten arabic forms
  using neural networks.}, in: \bibinfo{booktitle}{IEC (Prague)}, pp.
  \bibinfo{pages}{313--317}.
%Type = Article
\bibitem[{Awaidah and Mahmoud(2009)}]{awaidah2009multiple}
\bibinfo{author}{Awaidah, S.M.}, \bibinfo{author}{Mahmoud, S.A.},
  \bibinfo{year}{2009}.
\newblock \bibinfo{title}{A multiple feature/resolution scheme to arabic
  (indian) numerals recognition using hidden markov models}.
\newblock \bibinfo{journal}{Signal Processing} \bibinfo{volume}{89},
  \bibinfo{pages}{1176--1184}.
%Type = Inproceedings
\bibitem[{Awni et~al.(2019)Awni, Khalil and Abbas}]{awni2019deep}
\bibinfo{author}{Awni, M.}, \bibinfo{author}{Khalil, M.I.},
  \bibinfo{author}{Abbas, H.M.}, \bibinfo{year}{2019}.
\newblock \bibinfo{title}{Deep-learning ensemble for offline arabic handwritten
  words recognition}, in: \bibinfo{booktitle}{2019 14th International
  Conference on Computer Engineering and Systems (ICCES)},
  \bibinfo{organization}{IEEE}. pp. \bibinfo{pages}{40--45}.
%Type = Article
\bibitem[{Badry et~al.(2018)Badry, Hassan, Bayomi and Oakasha}]{badry2018qtid}
\bibinfo{author}{Badry, M.}, \bibinfo{author}{Hassan, H.},
  \bibinfo{author}{Bayomi, H.}, \bibinfo{author}{Oakasha, H.},
  \bibinfo{year}{2018}.
\newblock \bibinfo{title}{Qtid: Quran text image dataset}.
\newblock \bibinfo{journal}{International Journal Of Advanced Computer Science
  And Applications} \bibinfo{volume}{9}.
%Type = Article
\bibitem[{Badry et~al.(2021)Badry, Hassanin, Chandio and
  Moustafa}]{badry2021quranic}
\bibinfo{author}{Badry, M.}, \bibinfo{author}{Hassanin, M.},
  \bibinfo{author}{Chandio, A.}, \bibinfo{author}{Moustafa, N.},
  \bibinfo{year}{2021}.
\newblock \bibinfo{title}{Quranic script optical text recognition using deep
  learning in iot systems}.
\newblock \bibinfo{journal}{CMC-Comput. Mater. Contin} \bibinfo{volume}{68},
  \bibinfo{pages}{1847--1858}.
%Type = Article
\bibitem[{Bashir et~al.(2023)Bashir, Azmi, Nawaz, Zaghouani, Diab, Al-Fuqaha
  and Qadir}]{bashir2023arabic}
\bibinfo{author}{Bashir, M.H.}, \bibinfo{author}{Azmi, A.M.},
  \bibinfo{author}{Nawaz, H.}, \bibinfo{author}{Zaghouani, W.},
  \bibinfo{author}{Diab, M.}, \bibinfo{author}{Al-Fuqaha, A.},
  \bibinfo{author}{Qadir, J.}, \bibinfo{year}{2023}.
\newblock \bibinfo{title}{Arabic natural language processing for qur’anic
  research: A systematic review}.
\newblock \bibinfo{journal}{Artificial Intelligence Review}
  \bibinfo{volume}{56}, \bibinfo{pages}{6801--6854}.
%Type = Article
\bibitem[{Bassil and Alwani(2012)}]{bassil2012ocr}
\bibinfo{author}{Bassil, Y.}, \bibinfo{author}{Alwani, M.},
  \bibinfo{year}{2012}.
\newblock \bibinfo{title}{Ocr post-processing error correction algorithm using
  google online spelling suggestion}.
\newblock \bibinfo{journal}{arXiv preprint arXiv:1204.0191} .
%Type = Article
\bibitem[{Bataineh(2017)}]{bataineh2017printed}
\bibinfo{author}{Bataineh, B.}, \bibinfo{year}{2017}.
\newblock \bibinfo{title}{A printed paw image database of arabic language for
  document analysis and recognition.}
\newblock \bibinfo{journal}{Journal of ICT Research \& Applications}
  \bibinfo{volume}{11}.
%Type = Article
\bibitem[{Bergamaschi et~al.(2022)Bergamaschi, De~Nardis, Martoglia, Ruozzi,
  Sala, Vanzini and Vigliermo}]{bergamaschi2022novel}
\bibinfo{author}{Bergamaschi, S.}, \bibinfo{author}{De~Nardis, S.},
  \bibinfo{author}{Martoglia, R.}, \bibinfo{author}{Ruozzi, F.},
  \bibinfo{author}{Sala, L.}, \bibinfo{author}{Vanzini, M.},
  \bibinfo{author}{Vigliermo, R.A.}, \bibinfo{year}{2022}.
\newblock \bibinfo{title}{Novel perspectives for the management of multilingual
  and multialphabetic heritages through automatic knowledge extraction: The
  digitalmaktaba approach}.
\newblock \bibinfo{journal}{Sensors} \bibinfo{volume}{22},
  \bibinfo{pages}{3995}.
%Type = Article
\bibitem[{Bhatti et~al.(2023)Bhatti, Arif, Khalid, Khan, Ali, Khalid and
  Rehman}]{bhatti2023recognition}
\bibinfo{author}{Bhatti, A.}, \bibinfo{author}{Arif, A.},
  \bibinfo{author}{Khalid, W.}, \bibinfo{author}{Khan, B.},
  \bibinfo{author}{Ali, A.}, \bibinfo{author}{Khalid, S.},
  \bibinfo{author}{Rehman, A.u.}, \bibinfo{year}{2023}.
\newblock \bibinfo{title}{Recognition and classification of handwritten urdu
  numerals using deep learning techniques}.
\newblock \bibinfo{journal}{Applied Sciences} \bibinfo{volume}{13},
  \bibinfo{pages}{1624}.
%Type = Article
\bibitem[{Bilgin~Tasdemir(2023)}]{bilgin2023printed}
\bibinfo{author}{Bilgin~Tasdemir, E.F.}, \bibinfo{year}{2023}.
\newblock \bibinfo{title}{Printed ottoman text recognition using synthetic data
  and data augmentation}.
\newblock \bibinfo{journal}{International Journal on Document Analysis and
  Recognition (IJDAR)} , \bibinfo{pages}{1--15}.
%Type = Article
\bibitem[{Bin~Durayhim et~al.(2023)Bin~Durayhim, Al-Ajlan, Al-Turaiki and
  Altwaijry}]{bin2023towards}
\bibinfo{author}{Bin~Durayhim, A.}, \bibinfo{author}{Al-Ajlan, A.},
  \bibinfo{author}{Al-Turaiki, I.}, \bibinfo{author}{Altwaijry, N.},
  \bibinfo{year}{2023}.
\newblock \bibinfo{title}{Towards accurate children’s arabic handwriting
  recognition via deep learning}.
\newblock \bibinfo{journal}{Applied Sciences} \bibinfo{volume}{13},
  \bibinfo{pages}{1692}.
%Type = Article
\bibitem[{Boraik et~al.(2022)Boraik, Ravikumar and Saif}]{boraik2022characters}
\bibinfo{author}{Boraik, O.A.}, \bibinfo{author}{Ravikumar, M.},
  \bibinfo{author}{Saif, M.A.N.}, \bibinfo{year}{2022}.
\newblock \bibinfo{title}{Characters segmentation from arabic handwritten
  document images: Hybrid approach}.
\newblock \bibinfo{journal}{International Journal of Advanced Computer Science
  and Applications} \bibinfo{volume}{13}.
%Type = Inproceedings
\bibitem[{Boualam et~al.(2022)Boualam, Elfakir, Khaissidi and
  Mrabti}]{boualam2022arabic}
\bibinfo{author}{Boualam, M.}, \bibinfo{author}{Elfakir, Y.},
  \bibinfo{author}{Khaissidi, G.}, \bibinfo{author}{Mrabti, M.},
  \bibinfo{year}{2022}.
\newblock \bibinfo{title}{Arabic handwriting word recognition based on
  convolutional recurrent neural network}, in: \bibinfo{booktitle}{WITS 2020:
  Proceedings of the 6th International Conference on Wireless Technologies,
  Embedded, and Intelligent Systems}, \bibinfo{organization}{Springer}. pp.
  \bibinfo{pages}{877--885}.
%Type = Inproceedings
\bibitem[{Bouchakour et~al.(2021)Bouchakour, Meziani, Latrache, Ghribi and
  Yahiaoui}]{bouchakour2021printed}
\bibinfo{author}{Bouchakour, L.}, \bibinfo{author}{Meziani, F.},
  \bibinfo{author}{Latrache, H.}, \bibinfo{author}{Ghribi, K.},
  \bibinfo{author}{Yahiaoui, M.}, \bibinfo{year}{2021}.
\newblock \bibinfo{title}{Printed arabic characters recognition using combined
  features and cnn classifier}, in: \bibinfo{booktitle}{2021 International
  Conference on Recent Advances in Mathematics and Informatics (ICRAMI)},
  \bibinfo{organization}{IEEE}. pp. \bibinfo{pages}{1--5}.
%Type = Inproceedings
\bibitem[{Bouressace(2022)}]{bouressace2022review}
\bibinfo{author}{Bouressace, H.}, \bibinfo{year}{2022}.
\newblock \bibinfo{title}{A review of arabic document analysis methods}, in:
  \bibinfo{booktitle}{2022 4th International Conference on Pattern Analysis and
  Intelligent Systems (PAIS)}, \bibinfo{organization}{IEEE}. pp.
  \bibinfo{pages}{1--7}.
%Type = Article
\bibitem[{Butt et~al.(2021)Butt, Raza, Ramzan, Ali and
  Haris}]{butt2021attention}
\bibinfo{author}{Butt, H.}, \bibinfo{author}{Raza, M.R.},
  \bibinfo{author}{Ramzan, M.J.}, \bibinfo{author}{Ali, M.J.},
  \bibinfo{author}{Haris, M.}, \bibinfo{year}{2021}.
\newblock \bibinfo{title}{Attention-based cnn-rnn arabic text recognition from
  natural scene images}.
\newblock \bibinfo{journal}{Forecasting} \bibinfo{volume}{3},
  \bibinfo{pages}{520--540}.
%Type = Article
\bibitem[{Can and Kabaday{\i}(2020)}]{can2020automatic}
\bibinfo{author}{Can, Y.S.}, \bibinfo{author}{Kabaday{\i}, M.E.},
  \bibinfo{year}{2020}.
\newblock \bibinfo{title}{Automatic cnn-based arabic numeral spotting and
  handwritten digit recognition by using deep transfer learning in ottoman
  population registers}.
\newblock \bibinfo{journal}{Applied Sciences} \bibinfo{volume}{10},
  \bibinfo{pages}{5430}.
%Type = Inproceedings
\bibitem[{Chabchoub et~al.(2016)Chabchoub, Kessentini, Kanoun, Eglin and
  Lebourgeois}]{chabchoub2016smartatid}
\bibinfo{author}{Chabchoub, F.}, \bibinfo{author}{Kessentini, Y.},
  \bibinfo{author}{Kanoun, S.}, \bibinfo{author}{Eglin, V.},
  \bibinfo{author}{Lebourgeois, F.}, \bibinfo{year}{2016}.
\newblock \bibinfo{title}{Smartatid: A mobile captured arabic text images
  dataset for multi-purpose recognition tasks}, in: \bibinfo{booktitle}{2016
  15th International Conference on Frontiers in Handwriting Recognition
  (ICFHR)}, \bibinfo{organization}{IEEE}. pp. \bibinfo{pages}{120--125}.
%Type = Article
\bibitem[{Chorowski et~al.(2015)Chorowski, Bahdanau, Serdyuk, Cho and
  Bengio}]{chorowski2015attention}
\bibinfo{author}{Chorowski, J.}, \bibinfo{author}{Bahdanau, D.},
  \bibinfo{author}{Serdyuk, D.}, \bibinfo{author}{Cho, K.},
  \bibinfo{author}{Bengio, Y.}, \bibinfo{year}{2015}.
\newblock \bibinfo{title}{Attention-based models for speech recognition}.
\newblock \bibinfo{journal}{arXiv preprint arXiv:1506.07503} .
%Type = Article
\bibitem[{Djaghbellou et~al.(2021)Djaghbellou, Bouziane, Attia and
  Akhtar}]{djaghbellou2021survey}
\bibinfo{author}{Djaghbellou, S.}, \bibinfo{author}{Bouziane, A.},
  \bibinfo{author}{Attia, A.}, \bibinfo{author}{Akhtar, Z.},
  \bibinfo{year}{2021}.
\newblock \bibinfo{title}{A survey on arabic handwritten script recognition
  systems}.
\newblock \bibinfo{journal}{International Journal of Artificial Intelligence
  and Machine Learning (IJAIML)} \bibinfo{volume}{11}, \bibinfo{pages}{1--17}.
%Type = Article
\bibitem[{Doush et~al.(2018)Doush, Alkhateeb and Gharaibeh}]{doush2018novel}
\bibinfo{author}{Doush, I.A.}, \bibinfo{author}{Alkhateeb, F.},
  \bibinfo{author}{Gharaibeh, A.H.}, \bibinfo{year}{2018}.
\newblock \bibinfo{title}{A novel arabic ocr post-processing using rule-based
  and word context techniques}.
\newblock \bibinfo{journal}{International Journal on Document Analysis and
  Recognition (IJDAR)} \bibinfo{volume}{21}, \bibinfo{pages}{77--89}.
%Type = Inproceedings
\bibitem[{El-Sherif and Abdelazeem(2007)}]{el2007two}
\bibinfo{author}{El-Sherif, E.A.}, \bibinfo{author}{Abdelazeem, S.},
  \bibinfo{year}{2007}.
\newblock \bibinfo{title}{A two-stage system for arabic handwritten digit
  recognition tested on a new large database.}, in:
  \bibinfo{booktitle}{Artificial intelligence and pattern recognition}, pp.
  \bibinfo{pages}{237--242}.
%Type = Article
\bibitem[{Elkhayati et~al.(2022)Elkhayati, Elkettani and
  Mourchid}]{elkhayati2022segmentation}
\bibinfo{author}{Elkhayati, M.}, \bibinfo{author}{Elkettani, Y.},
  \bibinfo{author}{Mourchid, M.}, \bibinfo{year}{2022}.
\newblock \bibinfo{title}{Segmentation of handwritten arabic graphemes using a
  directed convolutional neural network and mathematical morphology
  operations}.
\newblock \bibinfo{journal}{Pattern Recognition} \bibinfo{volume}{122},
  \bibinfo{pages}{108288}.
%Type = Article
\bibitem[{Eltay et~al.(2020)Eltay, Zidouri and Ahmad}]{eltay2020exploring}
\bibinfo{author}{Eltay, M.}, \bibinfo{author}{Zidouri, A.},
  \bibinfo{author}{Ahmad, I.}, \bibinfo{year}{2020}.
\newblock \bibinfo{title}{Exploring deep learning approaches to recognize
  handwritten arabic texts}.
\newblock \bibinfo{journal}{IEEE Access} \bibinfo{volume}{8},
  \bibinfo{pages}{89882--89898}.
%Type = Article
\bibitem[{Eltay et~al.(2022)Eltay, Zidouri, Ahmad and
  Elarian}]{eltay2022generative}
\bibinfo{author}{Eltay, M.}, \bibinfo{author}{Zidouri, A.},
  \bibinfo{author}{Ahmad, I.}, \bibinfo{author}{Elarian, Y.},
  \bibinfo{year}{2022}.
\newblock \bibinfo{title}{Generative adversarial network based adaptive data
  augmentation for handwritten arabic text recognition}.
\newblock \bibinfo{journal}{PeerJ Computer Science} \bibinfo{volume}{8},
  \bibinfo{pages}{e861}.
%Type = Article
\bibitem[{Elzobi and Al-Hamadi(2018)}]{elzobi2018generative}
\bibinfo{author}{Elzobi, M.}, \bibinfo{author}{Al-Hamadi, A.},
  \bibinfo{year}{2018}.
\newblock \bibinfo{title}{Generative vs. discriminative recognition models for
  off-line arabic handwriting}.
\newblock \bibinfo{journal}{Sensors} \bibinfo{volume}{18},
  \bibinfo{pages}{2786}.
%Type = Article
\bibitem[{Farouk(2020)}]{mamdouh2020sentence}
\bibinfo{author}{Farouk, M.}, \bibinfo{year}{2020}.
\newblock \bibinfo{title}{Measuring text similarity based on structure and word
  embedding}.
\newblock \bibinfo{journal}{Cognitive Systems Research} \bibinfo{volume}{63}.
%Type = Incollection
\bibitem[{Frinken and Bunke(2014)}]{frinken2014continuous}
\bibinfo{author}{Frinken, V.}, \bibinfo{author}{Bunke, H.},
  \bibinfo{year}{2014}.
\newblock \bibinfo{title}{Continuous handwritten script recognition}, in:
  \bibinfo{editor}{Doermann, D.}, \bibinfo{editor}{Tombre, K.} (Eds.),
  \bibinfo{booktitle}{Handbook of Document Image Processing and Recognition}.
  \bibinfo{publisher}{Springer London}, \bibinfo{address}{London}, pp.
  \bibinfo{pages}{391--425}.
\newblock \DOIprefix\doi{10.1007/978-0-85729-859-1_12}.
%Type = Article
\bibitem[{Haghighi and Omranpour(2021)}]{haghighi2021stacking}
\bibinfo{author}{Haghighi, F.}, \bibinfo{author}{Omranpour, H.},
  \bibinfo{year}{2021}.
\newblock \bibinfo{title}{Stacking ensemble model of deep learning and its
  application to persian/arabic handwritten digits recognition}.
\newblock \bibinfo{journal}{Knowledge-Based Systems} \bibinfo{volume}{220},
  \bibinfo{pages}{106940}.
%Type = Article
\bibitem[{Hamad and Mehmet(2016)}]{hamad2016detailed}
\bibinfo{author}{Hamad, K.}, \bibinfo{author}{Mehmet, K.},
  \bibinfo{year}{2016}.
\newblock \bibinfo{title}{A detailed analysis of optical character recognition
  technology}.
\newblock \bibinfo{journal}{International Journal of Applied Mathematics
  Electronics and Computers} , \bibinfo{pages}{244--249}.
%Type = Inproceedings
\bibitem[{Hamada et~al.(2021)Hamada, Abdallah, Kasem and
  Abokhalil}]{hamada2021neural}
\bibinfo{author}{Hamada, M.A.}, \bibinfo{author}{Abdallah, A.},
  \bibinfo{author}{Kasem, M.}, \bibinfo{author}{Abokhalil, M.},
  \bibinfo{year}{2021}.
\newblock \bibinfo{title}{Neural network estimation model to optimize timing
  and schedule of software projects}, in: \bibinfo{booktitle}{2021 IEEE
  International Conference on Smart Information Systems and Technologies
  (SIST)}, \bibinfo{organization}{IEEE}. pp. \bibinfo{pages}{1--7}.
%Type = Article
\bibitem[{Hamida et~al.(2023)Hamida, Cherradi, El~Gannour, Raihani and
  Ouajji}]{hamida2023cursive}
\bibinfo{author}{Hamida, S.}, \bibinfo{author}{Cherradi, B.},
  \bibinfo{author}{El~Gannour, O.}, \bibinfo{author}{Raihani, A.},
  \bibinfo{author}{Ouajji, H.}, \bibinfo{year}{2023}.
\newblock \bibinfo{title}{Cursive arabic handwritten word recognition system
  using majority voting and k-nn for feature descriptor selection}.
\newblock \bibinfo{journal}{Multimedia Tools and Applications} ,
  \bibinfo{pages}{1--25}.
%Type = Inproceedings
\bibitem[{Huang et~al.(2019)Huang, Chen, He, Bai, Karatzas, Lu and
  Jawahar}]{huang2019icdar2019}
\bibinfo{author}{Huang, Z.}, \bibinfo{author}{Chen, K.}, \bibinfo{author}{He,
  J.}, \bibinfo{author}{Bai, X.}, \bibinfo{author}{Karatzas, D.},
  \bibinfo{author}{Lu, S.}, \bibinfo{author}{Jawahar, C.},
  \bibinfo{year}{2019}.
\newblock \bibinfo{title}{Icdar2019 competition on scanned receipt ocr and
  information extraction}, in: \bibinfo{booktitle}{2019 International
  Conference on Document Analysis and Recognition (ICDAR)},
  \bibinfo{organization}{IEEE}. pp. \bibinfo{pages}{1516--1520}.
%Type = Article
\bibitem[{Islam et~al.(2017)Islam, Islam and Noor}]{islam2017survey}
\bibinfo{author}{Islam, N.}, \bibinfo{author}{Islam, Z.},
  \bibinfo{author}{Noor, N.}, \bibinfo{year}{2017}.
\newblock \bibinfo{title}{A survey on optical character recognition system}.
\newblock \bibinfo{journal}{arXiv preprint arXiv:1710.05703} .
%Type = Article
\bibitem[{Jemni et~al.(2019)Jemni, Kessentini and Kanoun}]{jemni2019out}
\bibinfo{author}{Jemni, S.K.}, \bibinfo{author}{Kessentini, Y.},
  \bibinfo{author}{Kanoun, S.}, \bibinfo{year}{2019}.
\newblock \bibinfo{title}{Out of vocabulary word detection and recovery in
  arabic handwritten text recognition}.
\newblock \bibinfo{journal}{Pattern Recognition} \bibinfo{volume}{93},
  \bibinfo{pages}{507--520}.
%Type = Article
\bibitem[{Kasem et~al.(2022)Kasem, Abdallah, Berendeyev, Elkady, Abdalla,
  Mahmoud, Hamada, Nurseitov and Taj-Eddin}]{kasem2022deep}
\bibinfo{author}{Kasem, M.}, \bibinfo{author}{Abdallah, A.},
  \bibinfo{author}{Berendeyev, A.}, \bibinfo{author}{Elkady, E.},
  \bibinfo{author}{Abdalla, M.}, \bibinfo{author}{Mahmoud, M.},
  \bibinfo{author}{Hamada, M.}, \bibinfo{author}{Nurseitov, D.},
  \bibinfo{author}{Taj-Eddin, I.}, \bibinfo{year}{2022}.
\newblock \bibinfo{title}{Deep learning for table detection and structure
  recognition: A survey}.
\newblock \bibinfo{journal}{arXiv preprint arXiv:2211.08469} .
%Type = Article
\bibitem[{Khan and Adnan(2018)}]{khan2018urdu}
\bibinfo{author}{Khan, N.H.}, \bibinfo{author}{Adnan, A.},
  \bibinfo{year}{2018}.
\newblock \bibinfo{title}{Urdu optical character recognition systems: Present
  contributions and future directions}.
\newblock \bibinfo{journal}{IEEE Access} \bibinfo{volume}{6},
  \bibinfo{pages}{46019--46046}.
%Type = Article
\bibitem[{Khosravi and Kabir(2007)}]{khosravi2007introducing}
\bibinfo{author}{Khosravi, H.}, \bibinfo{author}{Kabir, E.},
  \bibinfo{year}{2007}.
\newblock \bibinfo{title}{Introducing a very large dataset of handwritten farsi
  digits and a study on their varieties}.
\newblock \bibinfo{journal}{Pattern recognition letters} \bibinfo{volume}{28},
  \bibinfo{pages}{1133--1141}.
%Type = Article
\bibitem[{Khosrobeigi et~al.(2022)Khosrobeigi, Veisi, Hoseinzade and
  Shabanian}]{khosrobeigi2022persian}
\bibinfo{author}{Khosrobeigi, Z.}, \bibinfo{author}{Veisi, H.},
  \bibinfo{author}{Hoseinzade, E.}, \bibinfo{author}{Shabanian, H.},
  \bibinfo{year}{2022}.
\newblock \bibinfo{title}{Persian optical character recognition using deep
  bidirectional long short-term memory}.
\newblock \bibinfo{journal}{Applied Sciences} \bibinfo{volume}{12},
  \bibinfo{pages}{11760}.
%Type = Inproceedings
\bibitem[{Lawgali et~al.(2013)Lawgali, Angelova and
  Bouridane}]{lawgali2013hacdb}
\bibinfo{author}{Lawgali, A.}, \bibinfo{author}{Angelova, M.},
  \bibinfo{author}{Bouridane, A.}, \bibinfo{year}{2013}.
\newblock \bibinfo{title}{Hacdb: Handwritten arabic characters database for
  automatic character recognition}, in: \bibinfo{booktitle}{European workshop
  on visual information processing (EUVIP)}, \bibinfo{organization}{IEEE}. pp.
  \bibinfo{pages}{255--259}.
%Type = Article
\bibitem[{Luqman et~al.(2014)Luqman, Mahmoud and Awaida}]{luqman2014kafd}
\bibinfo{author}{Luqman, H.}, \bibinfo{author}{Mahmoud, S.A.},
  \bibinfo{author}{Awaida, S.}, \bibinfo{year}{2014}.
\newblock \bibinfo{title}{Kafd arabic font database}.
\newblock \bibinfo{journal}{Pattern Recognition} \bibinfo{volume}{47},
  \bibinfo{pages}{2231--2240}.
%Type = Article
\bibitem[{Mahmoud and Kang(2023)}]{mahmoud2023ganmasker}
\bibinfo{author}{Mahmoud, M.}, \bibinfo{author}{Kang, H.S.},
  \bibinfo{year}{2023}.
\newblock \bibinfo{title}{Ganmasker: A two-stage generative adversarial network
  for high-quality face mask removal}.
\newblock \bibinfo{journal}{Sensors} \bibinfo{volume}{23},
  \bibinfo{pages}{7094}.
%Type = Inproceedings
\bibitem[{Mahmoud et~al.(2022)Mahmoud, Kasem, Abdallah and
  Kang}]{mahmoud2022ae}
\bibinfo{author}{Mahmoud, M.}, \bibinfo{author}{Kasem, M.},
  \bibinfo{author}{Abdallah, A.}, \bibinfo{author}{Kang, H.S.},
  \bibinfo{year}{2022}.
\newblock \bibinfo{title}{Ae-lstm: Autoencoder with lstm-based intrusion
  detection in iot}, in: \bibinfo{booktitle}{2022 International
  Telecommunications Conference (ITC-Egypt)}, \bibinfo{organization}{IEEE}. pp.
  \bibinfo{pages}{1--6}.
%Type = Article
\bibitem[{Mahmoud et~al.(2014)Mahmoud, Ahmad, Al-Khatib, Alshayeb, Parvez,
  M{\"a}rgner and Fink}]{mahmoud2014khatt}
\bibinfo{author}{Mahmoud, S.A.}, \bibinfo{author}{Ahmad, I.},
  \bibinfo{author}{Al-Khatib, W.G.}, \bibinfo{author}{Alshayeb, M.},
  \bibinfo{author}{Parvez, M.T.}, \bibinfo{author}{M{\"a}rgner, V.},
  \bibinfo{author}{Fink, G.A.}, \bibinfo{year}{2014}.
\newblock \bibinfo{title}{Khatt: An open arabic offline handwritten text
  database}.
\newblock \bibinfo{journal}{Pattern Recognition} \bibinfo{volume}{47},
  \bibinfo{pages}{1096--1112}.
%Type = Article
\bibitem[{Majumdar and Brick(2022)}]{majumdar2022recognizing}
\bibinfo{author}{Majumdar, S.}, \bibinfo{author}{Brick, A.},
  \bibinfo{year}{2022}.
\newblock \bibinfo{title}{Recognizing handwriting styles in a historical
  scanned document using scikit-fuzzy c-means clustering}.
\newblock \bibinfo{journal}{arXiv preprint arXiv:2210.16780} .
%Type = Article
\bibitem[{Malakar et~al.(2023)Malakar, Sahoo, Chakraborty, Sarkar and
  Nasipuri}]{malakar2023handwritten}
\bibinfo{author}{Malakar, S.}, \bibinfo{author}{Sahoo, S.},
  \bibinfo{author}{Chakraborty, A.}, \bibinfo{author}{Sarkar, R.},
  \bibinfo{author}{Nasipuri, M.}, \bibinfo{year}{2023}.
\newblock \bibinfo{title}{Handwritten arabic and roman word recognition using
  holistic approach}.
\newblock \bibinfo{journal}{The Visual Computer} \bibinfo{volume}{39},
  \bibinfo{pages}{2909--2932}.
%Type = Article
\bibitem[{Malhas and Elsayed(2022)}]{malhas2022arabic}
\bibinfo{author}{Malhas, R.}, \bibinfo{author}{Elsayed, T.},
  \bibinfo{year}{2022}.
\newblock \bibinfo{title}{Arabic machine reading comprehension on the holy
  qur’an using cl-arabert}.
\newblock \bibinfo{journal}{Information Processing \& Management}
  \bibinfo{volume}{59}, \bibinfo{pages}{103068}.
%Type = Article
\bibitem[{Mamoun(2023)}]{10.3103/S0146411623030069}
\bibinfo{author}{Mamoun, M.E.}, \bibinfo{year}{2023}.
\newblock \bibinfo{title}{An effective combination of convolutional neural
  network and support vector machine classifier for arabic handwritten
  recognition}.
\newblock \bibinfo{journal}{Autom. Control Comput. Sci.} \bibinfo{volume}{57},
  \bibinfo{pages}{267–275}.
\newblock \URLprefix \url{https://doi.org/10.3103/S0146411623030069},
  \DOIprefix\doi{10.3103/S0146411623030069}.
%Type = Inproceedings
\bibitem[{Milo and Martinez(2019)}]{milo2019new}
\bibinfo{author}{Milo, T.}, \bibinfo{author}{Martinez, A.G.},
  \bibinfo{year}{2019}.
\newblock \bibinfo{title}{A new strategy for arabic ocr: archigraphemes, letter
  blocks, script grammar, and shape synthesis}, in:
  \bibinfo{booktitle}{Proceedings of the 3rd International Conference on
  Digital Access to Textual Cultural Heritage}, pp. \bibinfo{pages}{93--96}.
%Type = Inproceedings
\bibitem[{Mohamed and Sayyed(2019)}]{mohamed2019arabic}
\bibinfo{author}{Mohamed, E.}, \bibinfo{author}{Sayyed, Z.A.},
  \bibinfo{year}{2019}.
\newblock \bibinfo{title}{Arabic-sos: segmentation, stemming, and orthography
  standardization for classical and pre-modern standard arabic}, in:
  \bibinfo{booktitle}{Proceedings of the 3rd International Conference on
  Digital Access to Textual Cultural Heritage}, pp. \bibinfo{pages}{27--32}.
%Type = Inproceedings
\bibitem[{Mortadi et~al.(2023)Mortadi, Mohamed, Talima, Alkhattip, Ibrahim,
  Osman and Hifny}]{mortadi2023alnasikh}
\bibinfo{author}{Mortadi, A.}, \bibinfo{author}{Mohamed, A.},
  \bibinfo{author}{Talima, A.}, \bibinfo{author}{Alkhattip, A.},
  \bibinfo{author}{Ibrahim, A.}, \bibinfo{author}{Osman, A.},
  \bibinfo{author}{Hifny, Y.}, \bibinfo{year}{2023}.
\newblock \bibinfo{title}{Alnasikh: An arabic ocr system based on
  transformers}, in: \bibinfo{booktitle}{2023 International Mobile,
  Intelligent, and Ubiquitous Computing Conference (MIUCC)},
  \bibinfo{organization}{IEEE}. pp. \bibinfo{pages}{74--81}.
%Type = Inproceedings
\bibitem[{Mostafa et~al.(2021)Mostafa, Mohamed, Ashraf, Elbehery, Jamal,
  Khoriba and Ghoneim}]{mostafa2021ocformer}
\bibinfo{author}{Mostafa, A.}, \bibinfo{author}{Mohamed, O.},
  \bibinfo{author}{Ashraf, A.}, \bibinfo{author}{Elbehery, A.},
  \bibinfo{author}{Jamal, S.}, \bibinfo{author}{Khoriba, G.},
  \bibinfo{author}{Ghoneim, A.S.}, \bibinfo{year}{2021}.
\newblock \bibinfo{title}{Ocformer: A transformer-based model for arabic
  handwritten text recognition}, in: \bibinfo{booktitle}{2021 International
  Mobile, Intelligent, and Ubiquitous Computing Conference (MIUCC)},
  \bibinfo{organization}{IEEE}. pp. \bibinfo{pages}{182--186}.
%Type = Article
\bibitem[{Mustapha et~al.(2022)Mustapha, Hasan, Nabus and
  Shamsuddin}]{mustapha2022conditional}
\bibinfo{author}{Mustapha, I.B.}, \bibinfo{author}{Hasan, S.},
  \bibinfo{author}{Nabus, H.}, \bibinfo{author}{Shamsuddin, S.M.},
  \bibinfo{year}{2022}.
\newblock \bibinfo{title}{Conditional deep convolutional generative adversarial
  networks for isolated handwritten arabic character generation}.
\newblock \bibinfo{journal}{Arabian Journal for Science and Engineering}
  \bibinfo{volume}{47}, \bibinfo{pages}{1309--1320}.
%Type = Article
\bibitem[{Nayef et~al.(2022)Nayef, Abdullah, Sulaiman and
  Alyasseri}]{nayef2022optimized}
\bibinfo{author}{Nayef, B.H.}, \bibinfo{author}{Abdullah, S.N.H.S.},
  \bibinfo{author}{Sulaiman, R.}, \bibinfo{author}{Alyasseri, Z.A.A.},
  \bibinfo{year}{2022}.
\newblock \bibinfo{title}{Optimized leaky relu for handwritten arabic character
  recognition using convolution neural networks}.
\newblock \bibinfo{journal}{Multimedia Tools and Applications} ,
  \bibinfo{pages}{1--30}.
%Type = Article
\bibitem[{Naz et~al.(2016)Naz, Umar, Shirazi, Ahmed, Razzak and
  Siddiqi}]{naz2016segmentation}
\bibinfo{author}{Naz, S.}, \bibinfo{author}{Umar, A.I.},
  \bibinfo{author}{Shirazi, S.H.}, \bibinfo{author}{Ahmed, S.B.},
  \bibinfo{author}{Razzak, M.I.}, \bibinfo{author}{Siddiqi, I.},
  \bibinfo{year}{2016}.
\newblock \bibinfo{title}{Segmentation techniques for recognition of
  arabic-like scripts: A comprehensive survey}.
\newblock \bibinfo{journal}{Education and Information Technologies}
  \bibinfo{volume}{21}, \bibinfo{pages}{1225--1241}.
%Type = Inproceedings
\bibitem[{Neudecker et~al.(2021)Neudecker, Baierer, Gerber, Clausner,
  Antonacopoulos and Pletschacher}]{neudecker2021survey}
\bibinfo{author}{Neudecker, C.}, \bibinfo{author}{Baierer, K.},
  \bibinfo{author}{Gerber, M.}, \bibinfo{author}{Clausner, C.},
  \bibinfo{author}{Antonacopoulos, A.}, \bibinfo{author}{Pletschacher, S.},
  \bibinfo{year}{2021}.
\newblock \bibinfo{title}{A survey of ocr evaluation tools and metrics}, in:
  \bibinfo{booktitle}{The 6th International Workshop on Historical Document
  Imaging and Processing}, pp. \bibinfo{pages}{13--18}.
%Type = Article
\bibitem[{Nguyen et~al.(2021)Nguyen, Jatowt, Coustaty and
  Doucet}]{nguyen2021survey}
\bibinfo{author}{Nguyen, T.T.H.}, \bibinfo{author}{Jatowt, A.},
  \bibinfo{author}{Coustaty, M.}, \bibinfo{author}{Doucet, A.},
  \bibinfo{year}{2021}.
\newblock \bibinfo{title}{Survey of post-ocr processing approaches}.
\newblock \bibinfo{journal}{ACM Computing Surveys (CSUR)} \bibinfo{volume}{54},
  \bibinfo{pages}{1--37}.
%Type = Article
\bibitem[{Nie et~al.(2015)Nie, Wang, Zhang, Yan, Zhang and
  Chua}]{nie2015disease}
\bibinfo{author}{Nie, L.}, \bibinfo{author}{Wang, M.}, \bibinfo{author}{Zhang,
  L.}, \bibinfo{author}{Yan, S.}, \bibinfo{author}{Zhang, B.},
  \bibinfo{author}{Chua, T.S.}, \bibinfo{year}{2015}.
\newblock \bibinfo{title}{Disease inference from health-related questions via
  sparse deep learning}.
\newblock \bibinfo{journal}{IEEE Transactions on knowledge and Data
  Engineering} \bibinfo{volume}{27}, \bibinfo{pages}{2107--2119}.
%Type = Article
\bibitem[{Nurseitov et~al.(2021a)Nurseitov, Bostanbekov, Kanatov, Alimova,
  Abdallah and Abdimanap}]{nurseitov2021classification}
\bibinfo{author}{Nurseitov, D.}, \bibinfo{author}{Bostanbekov, K.},
  \bibinfo{author}{Kanatov, M.}, \bibinfo{author}{Alimova, A.},
  \bibinfo{author}{Abdallah, A.}, \bibinfo{author}{Abdimanap, G.},
  \bibinfo{year}{2021}a.
\newblock \bibinfo{title}{Classification of handwritten names of cities and
  handwritten text recognition using various deep learning models}.
\newblock \bibinfo{journal}{arXiv preprint arXiv:2102.04816} .
%Type = Article
\bibitem[{Nurseitov et~al.(2021b)Nurseitov, Bostanbekov, Kurmankhojayev,
  Alimova, Abdallah and Tolegenov}]{nurseitov2021handwritten}
\bibinfo{author}{Nurseitov, D.}, \bibinfo{author}{Bostanbekov, K.},
  \bibinfo{author}{Kurmankhojayev, D.}, \bibinfo{author}{Alimova, A.},
  \bibinfo{author}{Abdallah, A.}, \bibinfo{author}{Tolegenov, R.},
  \bibinfo{year}{2021}b.
\newblock \bibinfo{title}{Handwritten kazakh and russian (hkr) database for
  text recognition}.
\newblock \bibinfo{journal}{Multimedia Tools and Applications}
  \bibinfo{volume}{80}, \bibinfo{pages}{33075--33097}.
%Type = Article
\bibitem[{Ouali et~al.(2023)Ouali, Halima and Wali}]{ouali2023augmented}
\bibinfo{author}{Ouali, I.}, \bibinfo{author}{Halima, M.B.},
  \bibinfo{author}{Wali, A.}, \bibinfo{year}{2023}.
\newblock \bibinfo{title}{An augmented reality for an arabic text reading and
  visualization assistant for the visually impaired}.
\newblock \bibinfo{journal}{Multimedia Tools and Applications} ,
  \bibinfo{pages}{1--29}.
%Type = Inproceedings
\bibitem[{Pechwitz et~al.(2002)Pechwitz, Maddouri, M{\"a}rgner, Ellouze, Amiri
  et~al.}]{pechwitz2002ifn}
\bibinfo{author}{Pechwitz, M.}, \bibinfo{author}{Maddouri, S.S.},
  \bibinfo{author}{M{\"a}rgner, V.}, \bibinfo{author}{Ellouze, N.},
  \bibinfo{author}{Amiri, H.}, et~al., \bibinfo{year}{2002}.
\newblock \bibinfo{title}{Ifn/enit-database of handwritten arabic words}, in:
  \bibinfo{booktitle}{Proc. of CIFED}, \bibinfo{organization}{Citeseer}. pp.
  \bibinfo{pages}{127--136}.
%Type = Inproceedings
\bibitem[{Prasad et~al.(2020)Prasad, Gadpal, Kapadni, Visave and
  Sultanpure}]{prasad2020cascadetabnet}
\bibinfo{author}{Prasad, D.}, \bibinfo{author}{Gadpal, A.},
  \bibinfo{author}{Kapadni, K.}, \bibinfo{author}{Visave, M.},
  \bibinfo{author}{Sultanpure, K.}, \bibinfo{year}{2020}.
\newblock \bibinfo{title}{Cascadetabnet: An approach for end to end table
  detection and structure recognition from image-based documents}, in:
  \bibinfo{booktitle}{Proceedings of the IEEE/CVF Conference on Computer Vision
  and Pattern Recognition Workshops}, pp. \bibinfo{pages}{572--573}.
%Type = Article
\bibitem[{Qaroush et~al.(2022a)Qaroush, Awad, Hanani, Mohammad, Jaber and
  Hasheesh}]{qaroush2022learning}
\bibinfo{author}{Qaroush, A.}, \bibinfo{author}{Awad, A.},
  \bibinfo{author}{Hanani, A.}, \bibinfo{author}{Mohammad, K.},
  \bibinfo{author}{Jaber, B.}, \bibinfo{author}{Hasheesh, A.},
  \bibinfo{year}{2022}a.
\newblock \bibinfo{title}{Learning-free, divide and conquer text-line
  extraction algorithm for printed arabic text with diacritics}.
\newblock \bibinfo{journal}{Journal of King Saud University-Computer and
  Information Sciences} \bibinfo{volume}{34}, \bibinfo{pages}{7699--7709}.
%Type = Article
\bibitem[{Qaroush et~al.(2022b)Qaroush, Jaber, Mohammad, Washaha, Maali and
  Nayef}]{qaroush2022efficient}
\bibinfo{author}{Qaroush, A.}, \bibinfo{author}{Jaber, B.},
  \bibinfo{author}{Mohammad, K.}, \bibinfo{author}{Washaha, M.},
  \bibinfo{author}{Maali, E.}, \bibinfo{author}{Nayef, N.},
  \bibinfo{year}{2022}b.
\newblock \bibinfo{title}{An efficient, font independent word and character
  segmentation algorithm for printed arabic text}.
\newblock \bibinfo{journal}{Journal of King Saud University-Computer and
  Information Sciences} \bibinfo{volume}{34}, \bibinfo{pages}{1330--1344}.
%Type = Article
\bibitem[{Ramdan et~al.(2013)Ramdan, Omar, Faidzul and Mady}]{ramdan2013arabic}
\bibinfo{author}{Ramdan, J.}, \bibinfo{author}{Omar, K.},
  \bibinfo{author}{Faidzul, M.}, \bibinfo{author}{Mady, A.},
  \bibinfo{year}{2013}.
\newblock \bibinfo{title}{Arabic handwriting data base for text recognition}.
\newblock \bibinfo{journal}{Procedia Technology} \bibinfo{volume}{11},
  \bibinfo{pages}{580--584}.
%Type = Inproceedings
\bibitem[{Rashid and Kumar~Gondhi(2022)}]{rashid2022scrutinization}
\bibinfo{author}{Rashid, D.}, \bibinfo{author}{Kumar~Gondhi, N.},
  \bibinfo{year}{2022}.
\newblock \bibinfo{title}{Scrutinization of urdu handwritten text recognition
  with machine learning approach}, in: \bibinfo{booktitle}{Emerging
  Technologies in Computer Engineering: Cognitive Computing and Intelligent
  IoT: 5th International Conference, ICETCE 2022, Jaipur, India, February 4--5,
  2022, Revised Selected Papers}, \bibinfo{organization}{Springer}. pp.
  \bibinfo{pages}{383--394}.
%Type = Article
\bibitem[{Reul et~al.(2019)Reul, Christ, Hartelt, Balbach, Wehner, Springmann,
  Wick, Grundig, B{\"u}ttner and Puppe}]{reul2019ocr4all}
\bibinfo{author}{Reul, C.}, \bibinfo{author}{Christ, D.},
  \bibinfo{author}{Hartelt, A.}, \bibinfo{author}{Balbach, N.},
  \bibinfo{author}{Wehner, M.}, \bibinfo{author}{Springmann, U.},
  \bibinfo{author}{Wick, C.}, \bibinfo{author}{Grundig, C.},
  \bibinfo{author}{B{\"u}ttner, A.}, \bibinfo{author}{Puppe, F.},
  \bibinfo{year}{2019}.
\newblock \bibinfo{title}{Ocr4all—an open-source tool providing a (semi-)
  automatic ocr workflow for historical printings}.
\newblock \bibinfo{journal}{Applied Sciences} \bibinfo{volume}{9},
  \bibinfo{pages}{4853}.
%Type = Inproceedings
\bibitem[{Sabbour and Shafait(2013)}]{sabbour2013segmentation}
\bibinfo{author}{Sabbour, N.}, \bibinfo{author}{Shafait, F.},
  \bibinfo{year}{2013}.
\newblock \bibinfo{title}{A segmentation-free approach to arabic and urdu ocr},
  in: \bibinfo{booktitle}{Document recognition and retrieval XX},
  \bibinfo{organization}{SPIE}. pp. \bibinfo{pages}{215--226}.
%Type = Inproceedings
\bibitem[{Sabri et~al.(2023)Sabri, Ennouni and Aarab}]{sabri2023robust}
\bibinfo{author}{Sabri, M.A.}, \bibinfo{author}{Ennouni, A.},
  \bibinfo{author}{Aarab, A.}, \bibinfo{year}{2023}.
\newblock \bibinfo{title}{A robust approach for arabic document images
  segmentation and indexation}, in: \bibinfo{booktitle}{International
  Conference on Digital Technologies and Applications},
  \bibinfo{organization}{Springer}. pp. \bibinfo{pages}{540--549}.
%Type = Inproceedings
\bibitem[{Saddami et~al.(2015)Saddami, Munadi and Arnia}]{saddami2015database}
\bibinfo{author}{Saddami, K.}, \bibinfo{author}{Munadi, K.},
  \bibinfo{author}{Arnia, F.}, \bibinfo{year}{2015}.
\newblock \bibinfo{title}{A database of printed jawi character image}, in:
  \bibinfo{booktitle}{2015 Third International Conference on Image Information
  Processing (ICIIP)}, \bibinfo{organization}{IEEE}. pp.
  \bibinfo{pages}{56--59}.
%Type = Inproceedings
\bibitem[{Salman and Altaei(2023)}]{salman2023proposed}
\bibinfo{author}{Salman, G.J.}, \bibinfo{author}{Altaei, M.S.M.},
  \bibinfo{year}{2023}.
\newblock \bibinfo{title}{Proposed deep learning system for arabic text
  detection and recognition}, in: \bibinfo{booktitle}{2023 15th International
  Conference on Developments in eSystems Engineering (DeSE)},
  \bibinfo{organization}{IEEE}. pp. \bibinfo{pages}{39--44}.
%Type = Article
\bibitem[{Singh et~al.(2012)Singh, Bacchuwar and Bhasin}]{singh2012survey}
\bibinfo{author}{Singh, A.}, \bibinfo{author}{Bacchuwar, K.},
  \bibinfo{author}{Bhasin, A.}, \bibinfo{year}{2012}.
\newblock \bibinfo{title}{A survey of ocr applications}.
\newblock \bibinfo{journal}{International Journal of Machine Learning and
  Computing} \bibinfo{volume}{2}, \bibinfo{pages}{314}.
%Type = Article
\bibitem[{Singh et~al.(2023)Singh, Garg and Kumar}]{singh2023performance}
\bibinfo{author}{Singh, S.}, \bibinfo{author}{Garg, N.K.},
  \bibinfo{author}{Kumar, M.}, \bibinfo{year}{2023}.
\newblock \bibinfo{title}{On the performance analysis of various features and
  classifiers for handwritten devanagari word recognition}.
\newblock \bibinfo{journal}{Neural Computing and Applications}
  \bibinfo{volume}{35}, \bibinfo{pages}{7509--7527}.
%Type = Article
\bibitem[{Slimane et~al.(2009)Slimane, Ingold, Kanoun, Alimi and
  Hennebert}]{slimane2009database}
\bibinfo{author}{Slimane, F.}, \bibinfo{author}{Ingold, R.},
  \bibinfo{author}{Kanoun, S.}, \bibinfo{author}{Alimi, A.M.},
  \bibinfo{author}{Hennebert, J.}, \bibinfo{year}{2009}.
\newblock \bibinfo{title}{Database and evaluation protocols for arabic printed
  text recognition}.
\newblock \bibinfo{journal}{DIUF-University of Fribourg-Switzerland}
  \bibinfo{volume}{1}.
%Type = Inproceedings
\bibitem[{Sulaiman et~al.(2017)Sulaiman, Omar and
  Nasrudin}]{sulaiman2017database}
\bibinfo{author}{Sulaiman, A.}, \bibinfo{author}{Omar, K.},
  \bibinfo{author}{Nasrudin, M.F.}, \bibinfo{year}{2017}.
\newblock \bibinfo{title}{A database for degraded arabic historical
  manuscripts}, in: \bibinfo{booktitle}{2017 6th International Conference on
  Electrical Engineering and Informatics (ICEEI)},
  \bibinfo{organization}{IEEE}. pp. \bibinfo{pages}{1--6}.
%Type = Inproceedings
\bibitem[{Tafti et~al.(2016)Tafti, Baghaie, Assefi, Arabnia, Yu and
  Peissig}]{tafti2016ocr}
\bibinfo{author}{Tafti, A.P.}, \bibinfo{author}{Baghaie, A.},
  \bibinfo{author}{Assefi, M.}, \bibinfo{author}{Arabnia, H.R.},
  \bibinfo{author}{Yu, Z.}, \bibinfo{author}{Peissig, P.},
  \bibinfo{year}{2016}.
\newblock \bibinfo{title}{Ocr as a service: an experimental evaluation of
  google docs ocr, tesseract, abbyy finereader, and transym}, in:
  \bibinfo{booktitle}{Advances in Visual Computing: 12th International
  Symposium, ISVC 2016, Las Vegas, NV, USA, December 12-14, 2016, Proceedings,
  Part I 12}, \bibinfo{organization}{Springer}. pp. \bibinfo{pages}{735--746}.
%Type = Article
\bibitem[{Tayyab et~al.(2022)Tayyab, Hussain, Alshara, Khan, Alotaibi and
  Baig}]{tayyab2022recognition}
\bibinfo{author}{Tayyab, M.}, \bibinfo{author}{Hussain, A.},
  \bibinfo{author}{Alshara, M.A.}, \bibinfo{author}{Khan, S.},
  \bibinfo{author}{Alotaibi, R.M.}, \bibinfo{author}{Baig, A.R.},
  \bibinfo{year}{2022}.
\newblock \bibinfo{title}{Recognition of visual arabic scripting news ticker
  from broadcast stream}.
\newblock \bibinfo{journal}{IEEE Access} \bibinfo{volume}{10},
  \bibinfo{pages}{59189--59204}.
%Type = Article
\bibitem[{Toiganbayeva et~al.(2022)Toiganbayeva, Kasem, Abdimanap, Bostanbekov,
  Abdallah, Alimova and Nurseitov}]{toiganbayeva2022kohtd}
\bibinfo{author}{Toiganbayeva, N.}, \bibinfo{author}{Kasem, M.},
  \bibinfo{author}{Abdimanap, G.}, \bibinfo{author}{Bostanbekov, K.},
  \bibinfo{author}{Abdallah, A.}, \bibinfo{author}{Alimova, A.},
  \bibinfo{author}{Nurseitov, D.}, \bibinfo{year}{2022}.
\newblock \bibinfo{title}{Kohtd: Kazakh offline handwritten text dataset}.
\newblock \bibinfo{journal}{Signal Processing: Image Communication}
  \bibinfo{volume}{108}, \bibinfo{pages}{116827}.
%Type = Article
\bibitem[{Waschneck et~al.(2018)Waschneck, Reichstaller, Belzner,
  Altenm{\"u}ller, Bauernhansl, Knapp and Kyek}]{waschneck2018optimization}
\bibinfo{author}{Waschneck, B.}, \bibinfo{author}{Reichstaller, A.},
  \bibinfo{author}{Belzner, L.}, \bibinfo{author}{Altenm{\"u}ller, T.},
  \bibinfo{author}{Bauernhansl, T.}, \bibinfo{author}{Knapp, A.},
  \bibinfo{author}{Kyek, A.}, \bibinfo{year}{2018}.
\newblock \bibinfo{title}{Optimization of global production scheduling with
  deep reinforcement learning}.
\newblock \bibinfo{journal}{Procedia Cirp} \bibinfo{volume}{72},
  \bibinfo{pages}{1264--1269}.
%Type = Article
\bibitem[{Xu et~al.(2021)Xu, Jang-Jaccard, Singh, Wei and
  Sabrina}]{xu2021improving}
\bibinfo{author}{Xu, W.}, \bibinfo{author}{Jang-Jaccard, J.},
  \bibinfo{author}{Singh, A.}, \bibinfo{author}{Wei, Y.},
  \bibinfo{author}{Sabrina, F.}, \bibinfo{year}{2021}.
\newblock \bibinfo{title}{Improving performance of autoencoder-based network
  anomaly detection on nsl-kdd dataset}.
\newblock \bibinfo{journal}{IEEE Access} \bibinfo{volume}{9},
  \bibinfo{pages}{140136--140146}.
%Type = Inproceedings
\bibitem[{Yousfi et~al.(2015)Yousfi, Berrani and Garcia}]{yousfi2015alif}
\bibinfo{author}{Yousfi, S.}, \bibinfo{author}{Berrani, S.A.},
  \bibinfo{author}{Garcia, C.}, \bibinfo{year}{2015}.
\newblock \bibinfo{title}{Alif: A dataset for arabic embedded text recognition
  in tv broadcast}, in: \bibinfo{booktitle}{2015 13th International Conference
  on Document Analysis and Recognition (ICDAR)}, \bibinfo{organization}{IEEE}.
  pp. \bibinfo{pages}{1221--1225}.
%Type = Inproceedings
\bibitem[{Zayene et~al.(2015)Zayene, Hennebert, Touj, Ingold and
  Amara}]{zayene2015dataset}
\bibinfo{author}{Zayene, O.}, \bibinfo{author}{Hennebert, J.},
  \bibinfo{author}{Touj, S.M.}, \bibinfo{author}{Ingold, R.},
  \bibinfo{author}{Amara, N.E.B.}, \bibinfo{year}{2015}.
\newblock \bibinfo{title}{A dataset for arabic text detection, tracking and
  recognition in news videos-activ}, in: \bibinfo{booktitle}{2015 13th
  International Conference on Document Analysis and Recognition (ICDAR)},
  \bibinfo{organization}{IEEE}. pp. \bibinfo{pages}{996--1000}.
%Type = Article
\bibitem[{Zayene et~al.(2018)Zayene, Masmoudi~Touj, Hennebert, Ingold and
  Essoukri Ben~Amara}]{zayene2018open}
\bibinfo{author}{Zayene, O.}, \bibinfo{author}{Masmoudi~Touj, S.},
  \bibinfo{author}{Hennebert, J.}, \bibinfo{author}{Ingold, R.},
  \bibinfo{author}{Essoukri Ben~Amara, N.}, \bibinfo{year}{2018}.
\newblock \bibinfo{title}{Open datasets and tools for arabic text detection and
  recognition in news video frames}.
\newblock \bibinfo{journal}{Journal of Imaging} \bibinfo{volume}{4},
  \bibinfo{pages}{32}.

\end{thebibliography}


\end{document}